\documentclass[times, 13pt, singlecolumn]{article} 
\usepackage{latex8}
\usepackage{listings}
\usepackage{times}
\usepackage{tikz}
\usepackage{placeins}
\usepackage{amssymb}
\usepackage{textgreek}
\usepackage{amsmath}
\usepackage[colorlinks,citecolor=green]{hyperref}
\usepackage{subfig}
\usepackage{algorithm}
\usepackage{algpseudocode}
\usepackage{booktabs}
\usepackage{graphicx}
\usetikzlibrary{positioning, arrows.meta}

\begin{document}

\title{Foundations and Models in Modern Computer Vision: Key Building Blocks in Landmark Architectures}

\author{
Radu-Andrei Bourceanu\thanks{These authors contributed equally to this report.}\\
radu.bourceanu@tum.de\\
\and
Neil De La Fuente\textsuperscript{*}\\
neil.de@tum.de\\
\and
Jan Grimm\textsuperscript{*}\\
jan.grimm@tum.de\\
\and
Andrei Jardan\textsuperscript{*}\\
andrei.jardan@tum.de\\
\and
Andriy Manucharyan\textsuperscript{*}\\
andriy.manucharyan@tum.de\\
\and
Cornelius Weiss\textsuperscript{*}\\
cornelius.weiss@tum.de\\
\and
Daniel Cremers\\
cremers@tum.de\\
\and
Roman Pflugfelder\\
roman.pflugfelder@tum.de\\
}

\maketitle
\thispagestyle{empty}

% in total 7 - 14 pages

%------------------------------------------------------------------------- 
% Machine learning with neural networks has become the prevalent view of visual learning in computer vision. Each year, the scientific community proposes dozens of new models as solutions for various visual tasks. A deeper understanding of the atomic components and building blocks that constitute a neural network is needed to better assess these models and compare them for a better understanding of their advantages and disadvantages.\\

% This seminar report attempts to distill such patterns of neural network models. We distinguish patterns arising at the level of network architecture and network layers. The architecture is a principled design of a neural network on a functional level, considering the possible inputs and outputs of a function and a functional classification. Generally, a neural network comprises several layers or sub-functions, considering a specific sub-step or task in the overall network architecture.\\

% The following examples were chosen from six students. In the first step, the students proposed and discussed several neural models and identified the different patterns and functions. In the second step, each student chose one paper and analysed its intent, the history of the work, the limitations, and the practical application. The results of these analyses are summarised in this report.

\begin{abstract}
This report analyzes the evolution of key design patterns in computer vision by examining six influential papers. The analysis begins with foundational architectures for image recognition. We review ResNet, which introduced residual connections to overcome the vanishing gradient problem and enable effective training of significantly deeper convolutional networks. Subsequently, we examine the Vision Transformer (ViT), which established a new paradigm by applying the Transformer architecture to sequences of image patches, demonstrating the efficacy of attention-based models for large-scale image recognition. Building on these visual representation backbones, we investigate generative models. Generative Adversarial Networks (GANs) are analyzed for their novel adversarial training process, which challenges a generator against a discriminator to learn complex data distributions. Then, Latent Diffusion Models (LDMs) are covered, which improve upon prior generative methods by performing a sequential denoising process in a perceptually compressed latent space. LDMs achieve high-fidelity synthesis with greater computational efficiency, representing the current state-of-the-art for image generation. Finally, we explore self-supervised learning techniques that reduce dependency on labeled data. DINO is a self-distillation framework in which a student network learns to match the output of a momentum-updated teacher, yielding features with strong k-NN classification performance. We conclude with Masked Autoencoders (MAE), which utilize an asymmetric encoder-decoder design to reconstruct heavily masked inputs, providing a highly scalable and effective method for pre-training large-scale vision models.
\end{abstract}

\newpage

\section{Introduction}
The field of computer vision has been defined by a series of architectural and methodological changes that have progressively advanced the capabilities of visual systems. This report provides a technical overview of six papers that represent critical milestones in this evolution, spanning foundational architectures, generative modeling, and self-supervised learning.

These six articles highlight three major phases in the trajectory of visual learning: (1) the development of strong feature extraction backbones, (2) the rise of generative modeling, and (3) the shift toward label-efficient training via self-supervised learning. In the first phase, deep learning for computer vision has progressed from convolutional networks like ResNet, which enabled deeper architectures by solving the vanishing gradient problem, to the Vision Transformer (ViT). ViTs replaced CNNs' traditional convolutional inductive biases with global self-attention on image patches to capture both local and long-range dependencies. The second phase introduced GANs and latent diffusion models that expand to generative tasks with distinct training dynamics and trade-offs between quality, diversity, and stability. Finally, the third phase explores how models like DINO and MAE learn robust representations without relying on labeled data, signaling the field's movement towards scalable and efficient visual understanding.

Together, these works represent key design patterns that continue to shape modern computer vision systems, from classification to image generation to self-supervised pre-training. This report aims to distill their core ideas and highlight their contributions to the evolving landscape of visual intelligence.

\section{Advances in Foundational Architectures}

Over the past decade, the field of deep learning has witnessed rapid development, with several key architectures marking certain points in the evolution of computer vision. These foundational architectures introduced new design principles and computational strategies that significantly advanced the capabilities of neural networks, enabling breakthroughs in image recognition, classification, segmentation, and more.
\\

In the following sections, two such architectures will be examined. First, we analyze Residual Network (ResNet), which introduced the concept of the residual block as a core building block, enabling the training of extremely deep networks by solving the vanishing gradient problem. 
The second part focuses on Visual Transformer Architecture (ViT), which brought the novel idea of applying transformer-based self-attention mechanisms, originally developed for natural language processing, directly to image data, redefining how visual information can be modeled without relying on convolutional layers.

\subsection{Residual Connections}
In the following section, we discuss the key contributions introduced by ResNet~\cite{he2016deep} (Deep Residual Learning for Image Recognition), focusing on the architecture of the residual block, the underlying intuition behind its design, and the empirical results demonstrating its effectiveness.
\subsubsection{Background and Motivation}
Before the introduction of residual learning, the training of very deep neural networks was limited by optimization difficulties such as vanishing gradients and degradation in performance. While deeper architectures were known to extract more abstract and complex features at different levels, empirical evidence showed that simply adding more layers often led to higher training error, contrary to expectation. Around 2012–2014, state-of-the-art convolutional neural networks like AlexNet~\cite{krizhevsky2012imagenet} (8 layers), VGGNet~\cite{Simonyan2014} (up to 19 layers), and GoogLeNet~\cite{szegedy2014goingdeeperconvolutions} (22 layers) demonstrated that increasing depth could improve accuracy, but only up to a certain point. Attempts to go deeper beyond 30 layers typically resulted in worse performance until the research team for ResNet introduced the ResNet architecture~\cite{aiblog}.
\\

Interestingly, the concept of additive information flow already existed in earlier models. Long Short-Term Memory networks (LSTMs), introduced in 1997~\cite{lstm}, addressed long-term dependencies in sequential data using additive memory cell updates, which allowed gradients to propagate through time more effectively. The forget gate in LSTMs functions similarly to a residual connection by allowing past information to flow directly into the current state. Similarly, Highway Networks (2015)~\cite{highway_networks} extended this idea to feedforward networks by introducing gated skip connections, allowing the network to learn when to carry forward or transform information. However, these mechanisms still relied on complex gating functions and were not widely adopted in vision tasks.
\\

The paper ``Deep Residual Learning for Image Recognition'' by He et al. (2015)~\cite{he2016deep} introduced a simple yet powerful solution: the residual connection. This identity shortcut adds the input of a layer to its output. This innovation enabled the successful training of networks with 50, 101, or even 152 layers, far beyond what was previously feasible. Residual Networks (ResNets) broke performance records on benchmarks like ImageNet and marked a fundamental shift in how deep architectures are designed. The residual framework provided a stable and efficient way to scale depth, effectively overcoming the optimization barriers that had previously constrained deep learning.

\subsubsection{Approach and Methodology}
\begin{figure}[H]
    \centering
    \includegraphics[width=0.85\textwidth]{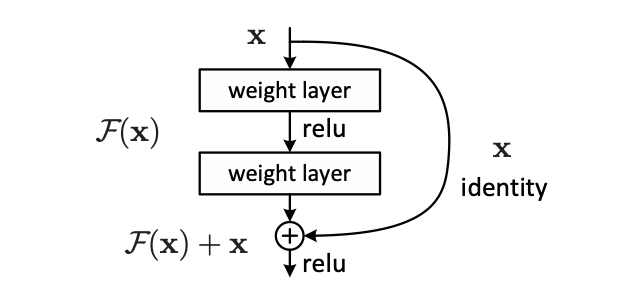}
    \caption{
    Residual learning: a building block. In courtesy of He et al.}
    \label{fig:res_learning}
\end{figure}

The core innovation of the ResNet architecture lies in the introduction of the residual block, a modular structure designed to simplify the training of very deep networks. Instead of learning a desired mapping directly, a residual block reformulates the problem: it learns the residual function 
$F(x) := H(x) - x$, where $H(x)$ is the target function and $x$ is the input (Figure~\ref{fig:res_learning}). This formulation leads to the final output being expressed as $H(x) = F(x) + x$, where the input $x$ is added via a shortcut (identity) connection to the output of a small stack of layers.

\begin{figure}[H]
    \centering
    \includegraphics[width=0.3\textwidth]{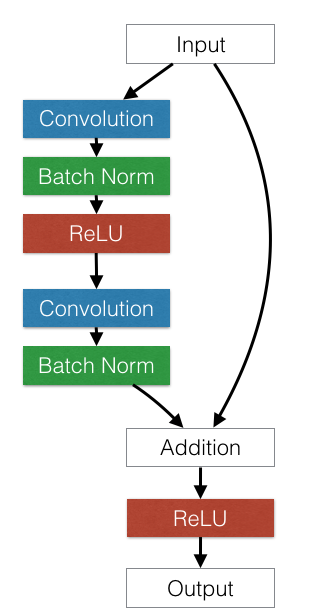}
    \caption{
    One variant of a residual block.}
    \label{fig:res_build}
\end{figure}

The intuition behind the residual block can be explained in two different ways:

\begin{enumerate}
    \item[1.] \textbf{Function approximation perspective:} Learning the residual function $F(x)$ is often easier than learning the entire transformation $H(x)$ from scratch. In many cases, the optimal transformation is close to the identity function. Thus, instead of reconstructing the full output from zero, the network only needs to learn the difference from the input to what has changed. This reduces the learning burden and helps the optimization process converge more easily.
    \item[2.] \textbf{Gradient flow perspective:} Residual connections also directly address the vanishing gradient problem, which previously limited the effective depth of deep networks. By allowing gradients to flow unimpeded through the identity paths, earlier layers receive stronger gradient signals during backpropagation. This preserves the ability of the network to adjust low-level representations even in very deep architectures.
\end{enumerate}
A standard residual block (Figure~\ref{fig:res_build}) consists of two or three convolutional layers, each followed by Batch Normalization (BN) and a ReLU activation. The shortcut path bypasses these layers (except for the last ReLU activation) and directly adds the input $x$ to the output of the final convolutional layer in the block. Importantly, the addition operation requires the dimensions of $x$ and $F(x)$ to match; if they do not, a linear projection (e.g., $1\times1$ convolution) is applied to $x$.

Over time, several variants of the residual block have been proposed and adopted in practice. For example, placing Batch Normalization after the addition, instead of after each convolution, has shown improved stability in some contexts. Additionally, some versions omit the final ReLU activation after the addition, allowing the identity mapping to pass through more directly.
\\

For their ImageNet submission, the authors introduced a more efficient variant called the bottleneck residual block. Instead of using two or three full convolution layers, the bottleneck design includes a $1\times1$ convolution for dimensionality reduction, followed by a $3\times3$ convolution for feature extraction, and another $1\times1$ convolution for dimensionality expansion. This design significantly reduces the number of parameters and computation, making it feasible to train much deeper networks (e.g., 50, 101, or 152 layers) without excessive memory or time costs.

\subsubsection{Experiments and Results}

\begin{figure}[H]
    \centering
    \includegraphics[width=0.85\textwidth]{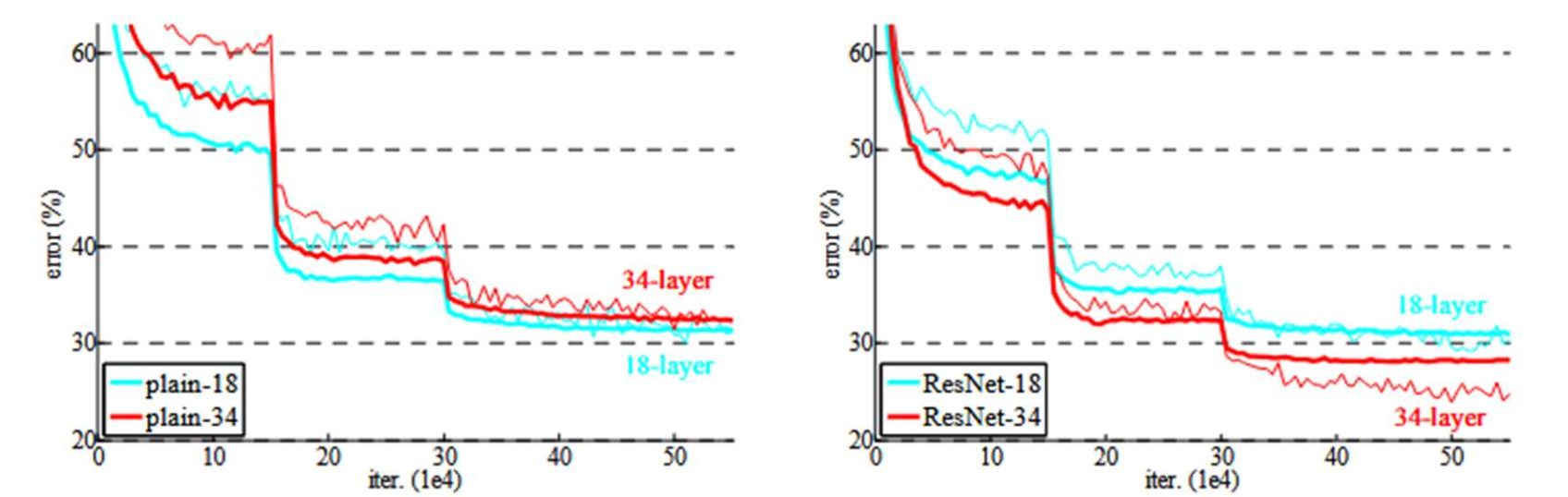}
    \caption{
    Training on ImageNet.}
    \label{fig:res_comparison}
\end{figure}

The authors of Deep Residual Learning for Image Recognition performed experiments to evaluate the effectiveness of residual learning on large-scale image classification tasks. A key finding was the impact of residual connections on network optimization, especially when increasing depth. In one experiment, a plain convolutional network with 34 layers was directly compared to a shallower 18-layer plain network, both using the same architectural design, which was inspired by VGGNet, but without residual connections (Figure~\ref{fig:res_comparison}). Thin curves denote training error, while bold curves denote validation error. 
\\

Surprisingly, the deeper 34-layer plain model had higher training loss and worse accuracy, demonstrating the degradation problem. However, when residual connections were introduced, the 34-layer ResNet significantly outperformed its 18-layer counterpart, both in training and test accuracy. This clearly showed that residual learning mitigates the degradation issue and enables the successful training of much deeper models.

\begin{figure}[H]
    \centering
    \includegraphics[width=0.5\textwidth]{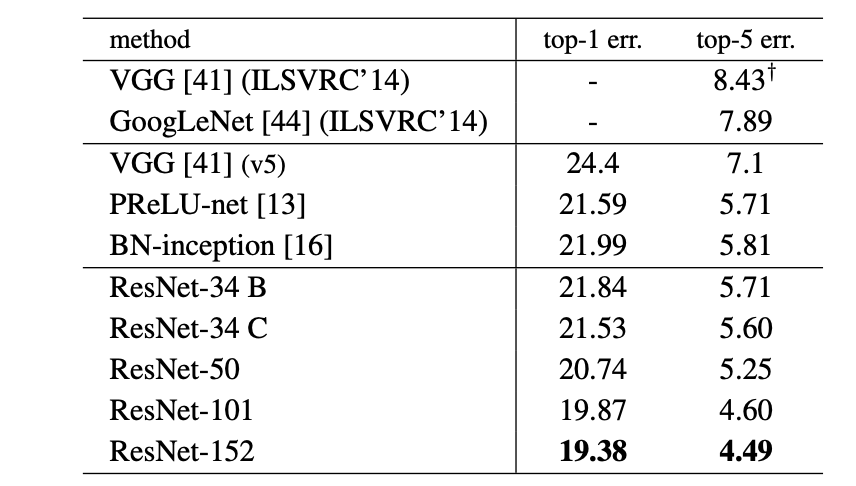}
    \caption{
    Error rates (\%) on the ImageNet validation ($\dagger$ - test) set.}
    \label{fig:resnet_table}
\end{figure}

In addition to qualitative improvements in training behavior, the ResNet architecture also achieved state-of-the-art performance on the ImageNet classification task. Deeper residual models ResNet-50, ResNet-101, and ResNet-152 consistently improved top-1 and top-5 accuracy on the validation set (Figure~\ref{fig:resnet_table}). ResNet-152 achieved a top-5 error of just 4.49\%, surpassing all previous models, including VGG-19 and GoogLeNet. These results validated the scalability of residual learning and highlighted the model’s capacity to generalize well, despite significantly increased depth.

\subsubsection{Discussion and Conclusion}
ResNet marked a turning point in deep learning by enabling the training of very deep convolutional networks without degradation, thanks to the concept of residual connections. Its success influenced the design of later models, and the residual block has since become a foundational component reused in a wide range of architectures. For example, Vision Transformers (ViTs)~\cite{dosovitskiy2020image} and other attention-based architectures have recently outperformed ResNets on many benchmarks, often with better generalization and less overfitting.
\\

Moreover, the residual learning principle has influenced domains beyond vision, including natural language processing, reinforcement learning, and audio modeling, highlighting its generality. In summary, ResNet laid the groundwork for modern deep architectures and continues to be a cornerstone in the evolution of deep learning.

\subsection{Vision Transformer}

The following subsection focuses on the Vision Transformer (ViT) that was introduced in the paper "An Image is Worth 16x16 Words: Transformers for Image Recognition at Scale" by Dosovitskiy et al. (2020)~\cite{dosovitskiy2020image}. First, the motivation behind Vision Transformers is explored, followed by an examination of their architecture and experimental results, and concluding with a short discussion.

\subsubsection{Background and Motivation}

Before the Vision Transformer was introduced, computer vision was dominated by Convolutional Neural Networks (CNNs). CNNs are specifically designed to process image data and are very effective at detecting patterns such as edges, textures, and shapes by using small filters that slide across the image. These filters focus on local regions, allowing CNNs to gradually learn increasingly complex features as more layers are added. CNNs have been highly successful in tasks like image classification, object detection, and image segmentation. Well-known models such as AlexNet~\cite{krizhevsky2012imagenet}, VGG~\cite{Simonyan2014}, and ResNet~\cite{he2016deep} have achieved impressive results across many vision benchmarks.

Meanwhile, in the field of natural language processing (NLP), the transformer architecture was gaining popularity. The original Transformer model, introduced in the paper “Attention is All You Need” by Vaswani et al. (2017)~\cite{vaswani_2017_attention}, showed that the self-attention mechanism could outperform traditional methods like Recurrent Neural Networks (RNNs) and Long Short-Term Memory (LSTM) networks. Transformers are particularly good at understanding relationships between words in a sentence, regardless of their position. They also allow for more parallelization during training, making them faster and more scalable. Models based on transformers, such as BERT~\cite{devlin-etal-2019-bert} and GPT~\cite{openai_chatgpt}, have become the foundation for many state-of-the-art NLP systems, performing well on tasks like translation, question answering, and text classification.
\\

The two distinct architectures, CNNs for computer vision and transformers for language processing, work very well on their specific tasks. So why even consider using transformers for vision tasks?

One major motivation is the idea of creating a universal architecture that can handle different types of input data, such as text and images, which would simplify the design of machine learning algorithms.

Another important aspect lies in how CNNs and transformers process data and the different inductive biases of the models. CNNs rely on convolutional layers with small, localized filters. These filters only consider information from neighboring pixels, meaning the model needs to stack many layers to eventually capture global relationships in the image. In contrast, transformers use self-attention mechanisms, which compute relationships between all input tokens at once, even in the first layer. This allows transformers to capture long-range dependencies and global context more effectively. In vision tasks, this can help the model understand broader spatial relationships in an image, not just local features. ~\cite{deininger2022comparative, tuli2021convolutional, vaswani_2017_attention}
\\

Initially, applying transformers to vision problems seemed counterintuitive, as transformers were originally designed for sequential data like text. However, researchers began experimenting with combining self-attention and convolution due to the advantages and potential of self-attention. For example, “Non-local Neural Networks” by Wang et al. (2018)~\cite{wang2018non} introduced a hybrid approach where only a few self-attention layers were added to a CNN architecture. These attention layers allowed each pixel to gather information from distant parts of the image, improving the model’s ability to understand global patterns. Still, this approach mostly relied on convolutional operations and only partially integrated self-attention.

The Vision Transformer takes a more direct approach; it removes convolutions entirely, proving that the reliance on CNNs is not necessary. ViT demonstrates that pure transformer architectures can perform very well on vision tasks, sometimes even outperforming CNNs~\cite{dosovitskiy2020image}.

\subsubsection{Approach and Methodology}

To understand how the Vision Transformer (ViT) works for image classification, it's helpful to first look at how transformer models are used in text classification, since they follow a very similar approach.
For example, BERT, introduced in the paper "BERT: Pre-training of deep bidirectional transformers for language understanding" by Devlin et al. (2019)~\cite{devlin-etal-2019-bert}, is a well-known model when it comes to text classification. BERT takes a sequence of words and predicts a class label for the input text. It can be used for different classification tasks like sentiment analysis, where the goal is to classify the tone of a sentence or document. For example, BERT can be trained to label movie reviews as positive, negative, or neutral, to evaluate how well a movie is received by the audience~\cite{alaparthi2021bert}. Another task might be emotion classification, where the model learns to label text as funny, sad, or angry. Since classification tasks require only a single output, BERT's architecture consists only of transformer encoders. These encoders are used to extract the content and meaning of the input text and store all the information in a single classification token. This token is then used to predict the most likely class labels.

This is conceptually quite similar to image classification, where the input is an image and the model must predict a label such as "bird," "car," or "house." The key challenge lies in adapting the image data so it can be processed by a transformer, which expects a sequence of tokens as input.

With text, this transformation is straightforward; the input sentences are split into tokens, and each token is embedded into a high-dimensional vector space. Positional encodings are then added to preserve the order of the tokens, since transformers do not have any inherent understanding of sequence or position. The resulting sequence of vectors is then passed into the transformer encoder.

\begin{figure}[H]
    \centering
    \includegraphics[width=0.85\textwidth]{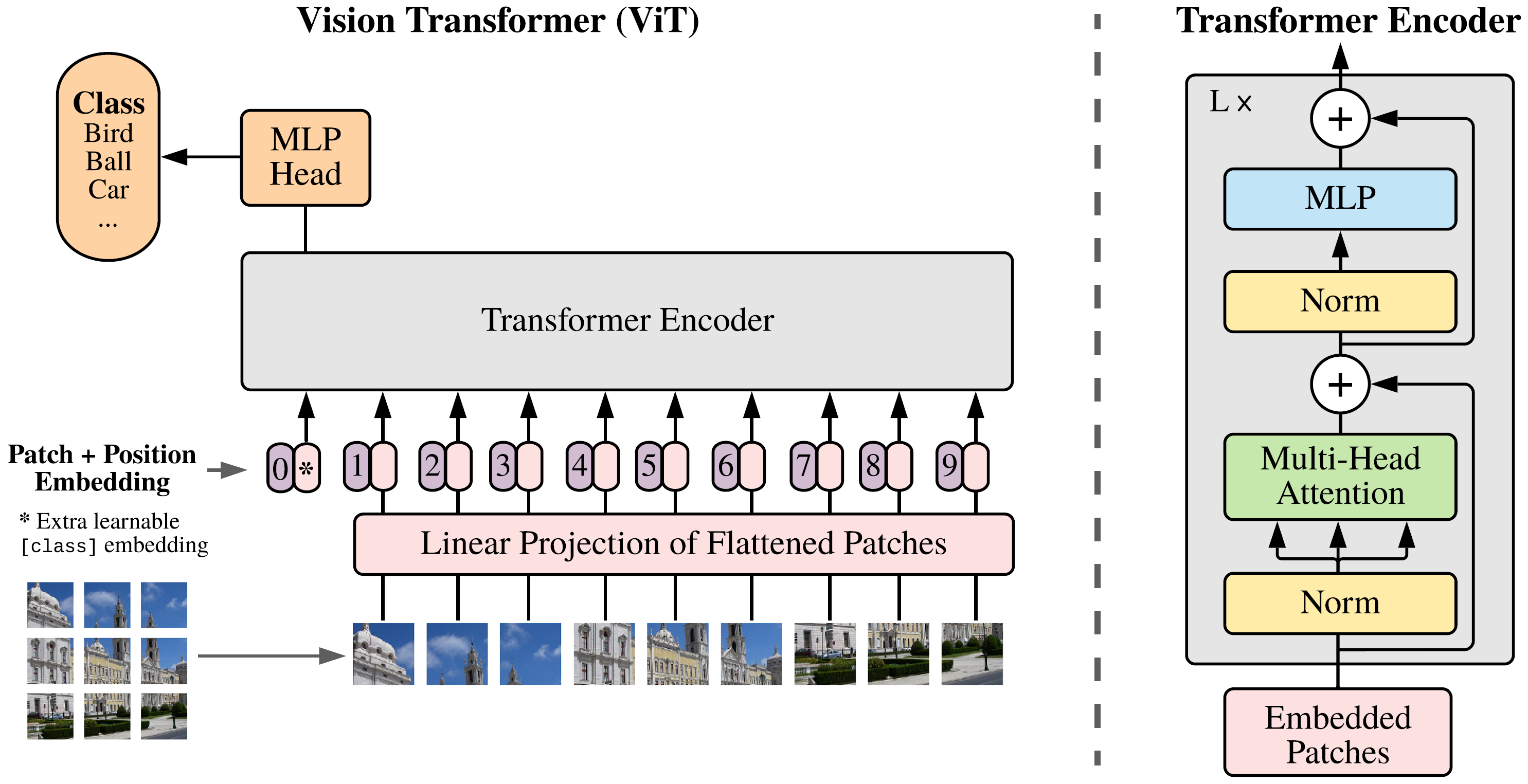}
    \caption{
    Vision Transformer Architecture~\cite{dosovitskiy2020image}. In courtesy of the authors.}
    \label{fig:vit}
\end{figure}

To apply this idea to images, the image must first be sequentialized so that it can be accepted by the transformer encoder. This is done by dividing the input image into small, fixed-size patches, such as 16x16 pixels. Each patch is then flattened and passed through a linear projection layer to map it into the embedding space. After this step, positional embeddings are added, and finally, all embeddings plus the classification token are fed into the encoder. This CLS token acts as a summary representation, similar to how it is used in BERT. The entire sequence is then passed through multiple standard transformer encoder layers, where self-attention mechanisms allow each patch to interact with all others. This enables the model to capture both local and global relationships within the image right from the start (Figure~\ref{fig:vit}).

After processing through the transformer, the output corresponding to the classification token, which now contains information about the entire input, is passed to a final classification layer to predict the image’s label.

This approach is interesting because it completely avoids using convolutional layers. Instead, the model learns to interpret the full image using only self-attention and feedforward layers. This enables the Vision Transformer to capture global structures early in the model, offering a fundamentally different approach.

\subsubsection{Experiments and Results}

In the original Vision Transformer (ViT) paper, the authors focused their performance evaluations on image classification tasks~\cite{dosovitskiy2020image}. Since transformers typically require large amounts of data to perform well, ViT was first pre-trained on large-scale datasets and then fine-tuned on smaller downstream tasks, following a common transfer learning approach.
\\

Multiple versions of the Vision Transformer were tested, differing mainly in the number of layers, size of the embedding dimension, and MLP hidden size. For example, the model ViT-Huge consists of 32 encoder layers, has an embedding size of 1280, and the MLP has a dimension of 5120. Thus, ViT-H/14 stands for the ViT-Huge model, and images were split into 14x14 patches. These were compared against the ResNet-based BiT~\cite{kolesnikov2020big} (Big Transfer) model, which was at the time one of the top-performing CNN architectures for transfer learning in vision tasks.

\begin{figure}[H]
    \centering
    \includegraphics[width=1.0\textwidth]{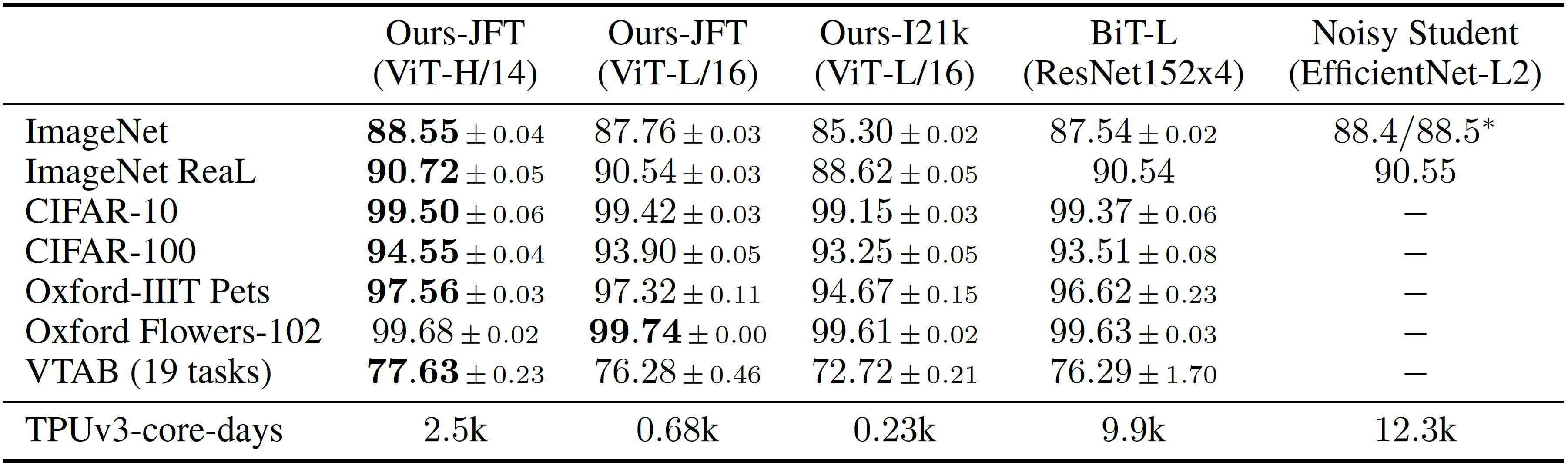}
    \caption{
    Comparison of Vision Transformer and CNNs on multiple image classification datasets.~\cite{dosovitskiy2020image}}
    \label{fig:vit_results}
\end{figure}

The ViT-H outperforms ResNet across all evaluated datasets (Figure~\ref{fig:vit_results}). Despite the lack of convolutional layers, ViT achieves higher accuracy and, at the same time, requires fewer computational resources to train.

\begin{figure}[H]
    \centering
    \includegraphics[width=0.60\textwidth]{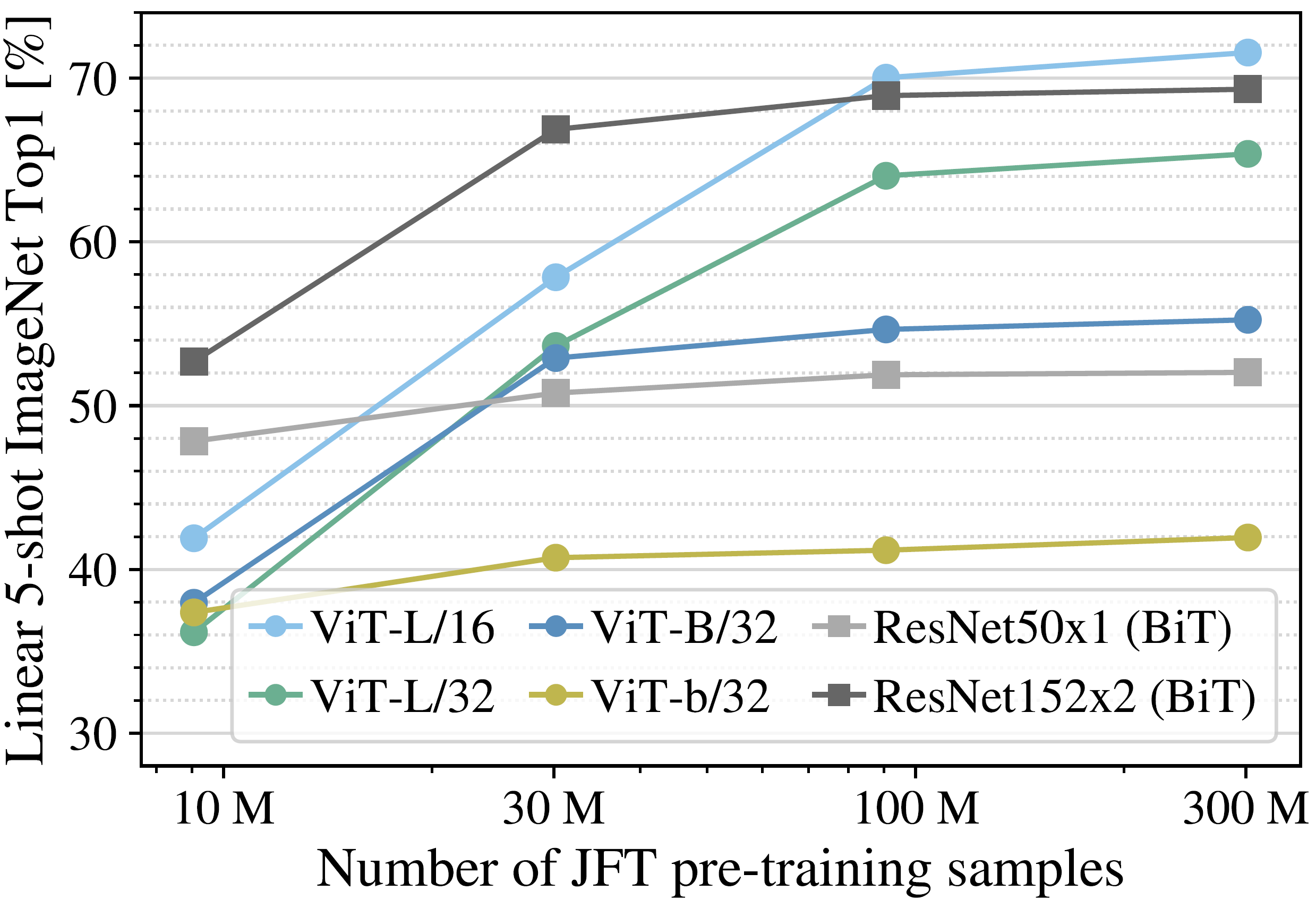}
    \caption{
    Comparison of Vision Transformer and ResNet on different training dataset sizes.~\cite{dosovitskiy2020image}}
    \label{fig:vit_results2}
\end{figure}

Another graph offers a deeper look into how ViT performs depending on the amount of data available during pre-training (Figure~\ref{fig:vit_results2}). It shows that when the model is pre-trained on smaller datasets, the ResNet (BiT) still performs better. However, as the size of the pre-training dataset increases, the Vision Transformer starts to outperform ResNet.
\\

This result can be explained by the differences in architecture. CNNs generally have more inductive bias. They have strong built-in assumptions about the structure of images, such as locality and translation invariance, thanks to their use of convolutional filters. These assumptions help CNNs generalize well from smaller datasets because they guide the learning process toward patterns that are usually relevant in images.

Transformers, on the other hand, have less inductive bias. They do not assume locality or spatial structure and instead rely on the model to learn all relevant patterns from the data using self-attention. It also means that they tend to overfit with small datasets. When enough data is available, ViT is more flexible and can discover and leverage these patterns on its own, leading to better performance than CNNs.~\cite{vaswani_2017_attention}
\\

In summary, the experiments show that Vision Transformers can outperform state-of-the-art CNNs when sufficient training data is available. This result demonstrates the potential of transformer-based models as a general-purpose solution for vision tasks, especially in settings where large-scale data is accessible.

\subsubsection{Discussion and Conclusion}

The Vision Transformer represents a major shift in how we approach computer vision tasks. Traditionally, Convolutional Neural Networks have dominated this field due to their strong inductive biases. These properties allow CNNs to perform well even with limited data, as they are designed to exploit the spatial structure of images through convolutional filters.
\\

Transformers, originally developed for natural language processing, take a fundamentally different approach. Instead of using convolutions, they rely on self-attention mechanisms that allow each input element to attend to every other element. This global attention mechanism enables the model to capture long-range dependencies and complex relationships in the input data right from the first layer.

The key idea behind Vision Transformers is to treat an image as a sequence of patches, similar to how text is treated as a sequence of words or tokens. By embedding these patches and feeding them into a standard transformer encoder, the model learns to recognize visual patterns without any convolutional operations. This approach makes use of less inductive bias and lets the model learn directly from the data.
\\

Experiments have shown that ViT can outperform traditional CNNs like ResNet when pre-trained on sufficiently large datasets. While CNNs still perform better with sparse data, transformers become more effective when more data is available.

In addition to image classification, the ViT architecture has also proven to be highly adaptable. Variants of the Vision Transformer have been successfully used for tasks such as object detection~\cite{liu2021swin} and self-supervised learning~\cite{caron2021emerging}. This flexibility makes transformers a promising general-purpose backbone for a wide range of vision applications.
\\

In summary, the Vision Transformer shows that convolution is not strictly necessary for solving vision tasks. As larger vision datasets might become available and training gets better in the future, transformer-based models are expected to become more important in computer vision~\cite{zhai2022scaling}.

\section{Paradigms in Generative Modeling}
%A little intro here would be nice (Cornelius)

Generative tasks in ML focus on creating new data samples that resemble a given input data set. Unlike discriminative tasks, which aim to classify or segment input data, generative tasks attempt to model the underlying data distribution to produce new plausible data points. In computer vision, this involves generating realistic and consistent images, videos, or 3D objects. A well-trained generative model might, for example, up-sample the resolution of existing images (image enhancement), synthesize multiple images (image synthesis), create new images from text (text-to-image generation), or translate sketches into photographs (image-to-image translation). These tasks are more challenging, because they require the model to understand both the global structure and the fine-grained details of the visual data. In addition, models must learn to simulate the physics and behavior of objects and living beings. Human movements, the consistent use and position of fingers, or the generation of realistic smoke are considered difficult~\cite{shamsolmoali2020imagesynthesisadversarialnetworks}. In the following sections, two different approaches to image generation will be examined. First, Generative Adversarial Networks (GANs) are discussed, known for their unstable training process, involving two separate models. The second part focuses on Latent Diffusion Models (LDMs), which are widely used because of their stable training and realistic results.

\subsection{Generative Adversarial Networks}

In the following subsection, the architecture, training dynamics and results of Generative Adversarial Networks (GANs) are discussed, as originally proposed by Goodfellow et al.~\cite{goodfellow2014generative}. In addition, several architectural and methodological advances are reviewed. In particular, Progressive Growing of GANs~\cite{karras2018progressivegrowinggansimproved}, Progressive Augmentation GAN~\cite{zhang2019progressiveaugmentationgans}, Energy-based GANs~\cite{zhao2017energybasedgenerativeadversarialnetwork} and BigGAN~\cite{brock2019largescalegantraining} aim to mitigate key challenges such as training instability, generator-discriminator desynchronization and approximation of the underlying data distribution.

\subsubsection{Background and Motivation}

Before GANs, advances in computer vision were dominated by discriminative models, which are based on convolutions. In contrast, generative models, which aim to produce new samples by learning the input data distribution, faced significant hurdles. Key challenges include approximating probability distributions, reliance on computationally expensive methods (e.g. Markov Chain Monte Carlo) and limited diversity in generated samples.

In 2014, Goodfellow et al. introduced Generative Adversarial Networks (GANs) to overcome these limitations. Instead of explicitly modeling the data distribution, they framed generation as a game between two competing networks. The generator (\( G \)) learns to recreate data from random noise, while the discriminator (\( D \)) tries to distinguish real images from generations. The GAN framework allows direct training via backpropagation, eliminating the need for computationally expensive Markov Chains and inference. Furthermore, the framework is flexible, accommodating diverse architectures for (\( G \)) and (\( D \)). GANs allowed for sharp, high-quality output that differentiated them from previous models. However, they also posed new challenges like training instability, \textit{mode collapse} and the need to balance both networks. These challenges have inspired many follow-up research articles focused on stabilizing training.

\subsubsection{Approach and Methodology}

Generative Adversarial Networks propose a game-theoretic framework formed by two competing neural networks. The generator \( G(z; \theta_g) \) maps random noise vectors \( z \sim p_z(z) \) to the data space, while the discriminator \( D(x; \theta_d) \) outputs a scalar representing the probability that the input \( x \) comes from the real data distribution \( p_{\text{data}}(x) \) rather than from \( G \). The training objective is formulated as a two-player min/max game:

\[
\min_G \max_D V(D, G) = \mathbb{E}_{x \sim p_{\text{data}}}[\log D(x)] + \mathbb{E}_{z \sim p_z}[\log(1 - D(G(z)))]
\]

The discriminator is trained to (a) classify real samples as real by maximizing \( \log D(x) \) and (b) label generated images as fake by maximizing \( \log (1 - D(G(z))) \). Moreover, the generator is trained to minimize \( \log (1 - D(G(z))) \), thus 'fool' the discriminator into thinking that the generated images are real.

In practice, both \( G \) and \( D \) are implemented as multilayer perceptrons (MLP), trained by backpropagation and dropout. Addressing vanishing gradients \( G \) is maximizing \( \log D(G(z)) \) instead, which provides the same fix-point dynamic for \( G \) and \( D \), but offers much stronger gradients early on.

Training proceeds by alternating \( k \) steps for updating the generator and one step for the discriminator. This ensures that \( D \) remains close to optimal as \( G \) gradually improves. Training \( G \) and \( D \) in the same loop leads to bad gradient updates for \( G \) and overfitting in \( D \). No Markov chains or inference are required, only gradient-based updates, improving the overall computational efficiency.

\subsubsection{Experiments}

Trained and evaluated on the following benchmark datasets, GANs demonstrate the ability to learn complex data distributions through adversarial training:

\begin{itemize}
    \item The \textbf{MNIST digits dataset}, consisting of 28$\times$28 px grayscale images of handwritten digits.
    \item The \textbf{Toronto Face Dataset (TFD)}, composed of 48$\times$48 px grayscale facial images.
    \item The \textbf{CIFAR-10 dataset}, consisting of 32$\times$32 px color images across 10 object categories.
\end{itemize}

\begin{figure}[h]
    \centering
    \includegraphics[width=0.85\textwidth]{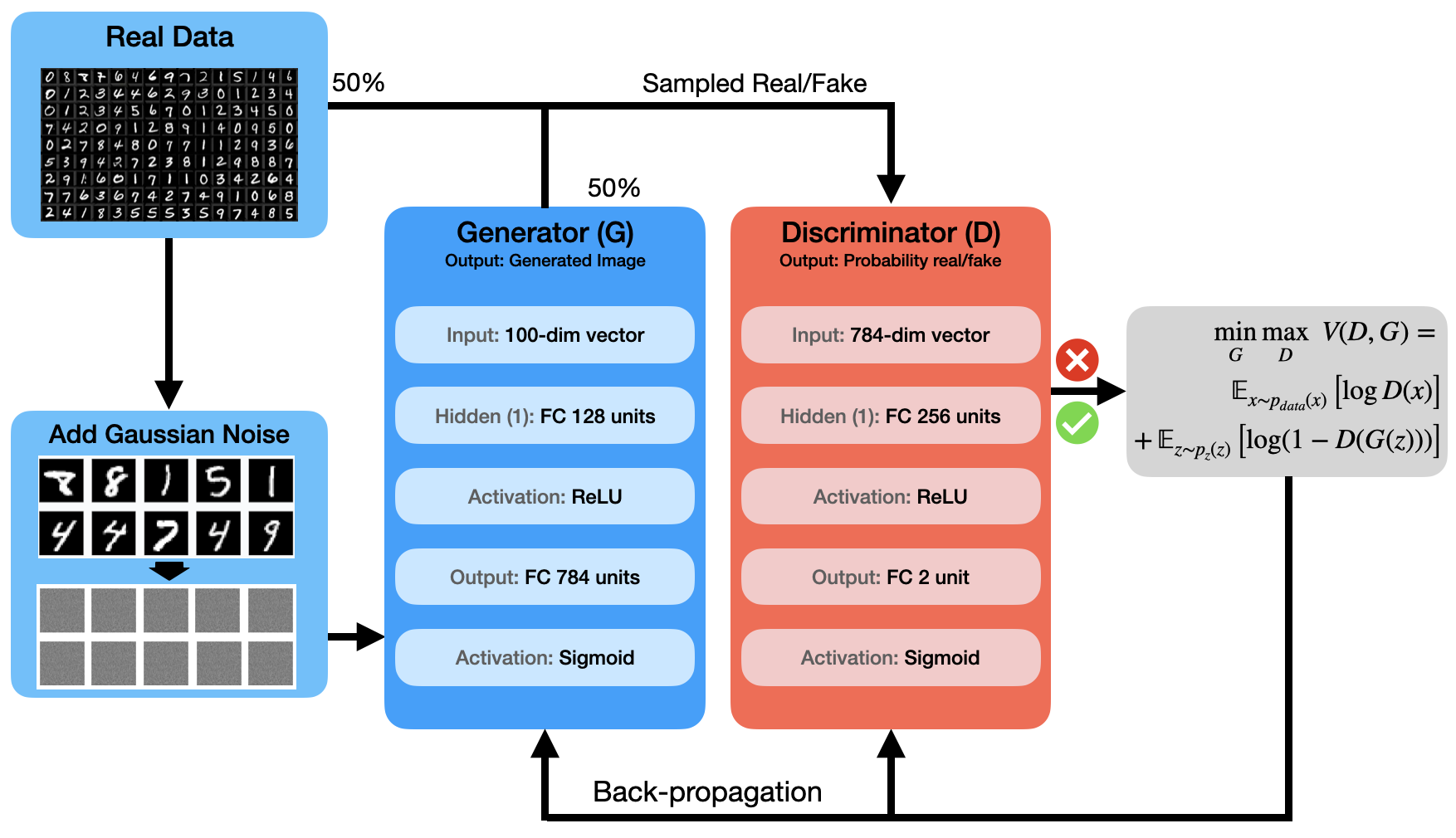}
    \caption{
    Overview of the vanilla GAN framework. 
    \label{fig:gan_flow_diagram}
    }
\end{figure}

The generator receives a noise vector \( z \), sampled from a uniform or Gaussian distribution and maps it to a synthetic image \( G(z) \). Generated samples are shuffled with real images in a 50/50 ratio and passed to the discriminator. \( D \) classifies them (real/fake) and outputs a probability score \( D(x) \). This score is used to compute the adversarial loss \( V(D, G) \), which is minimized with respect to the generator and maximized with respect to the discriminator. Gradients from loss are propagated back to update the weights of both MLPs (Figure~\ref{fig:gan_flow_diagram}). Vanilla GANs are trained using an ADAM optimizer with a standard learning rate of 0.001. 

\begin{figure}[h]
    \centering
    \includegraphics[width=0.85\textwidth]{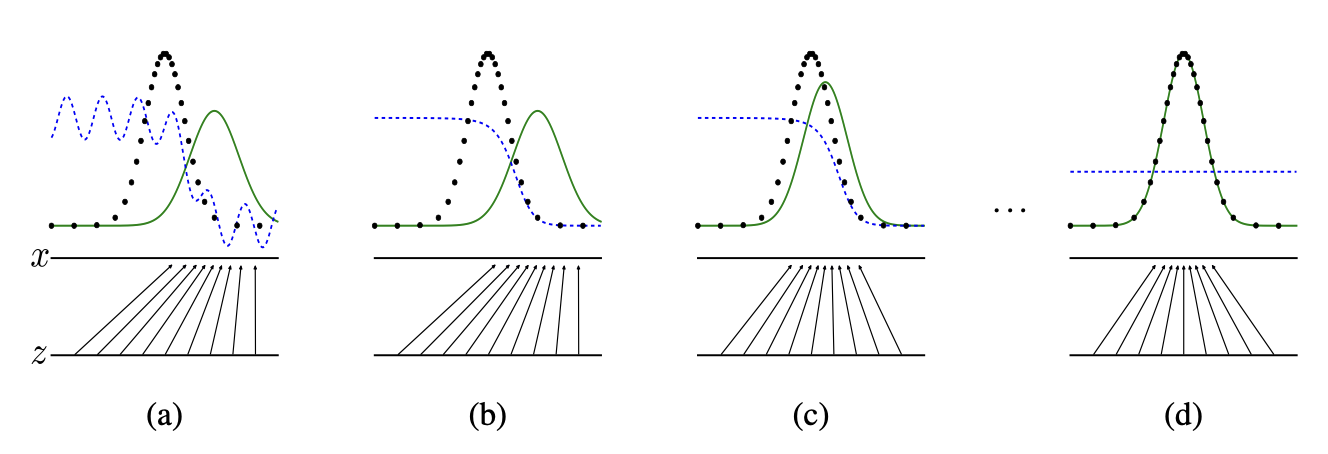}
    \caption{
    Visualization of distributions during GAN training. 
    \textbf{Black dotted line}: True data distribution \( p_{\text{data}} \); 
    \textbf{Green solid line}: Generator distribution \( p_g \); 
    \textbf{Blue dashed line}: Discriminator output \( D(x) \). 
    (a) Initial state: Randomly initialized generator and discriminator. 
    (b) Discriminator learns to distinguish real and fake data. 
    (c) Generator improves; fake samples get closer to real data. 
    (d) Convergence: Generator matches the real data distribution; discriminator outputs 0.5 everywhere. 
    Adapted from~\cite{goodfellow2014generative}.}
    \label{fig:training_distributions}
\end{figure}

To illustrate training dynamics, the authors visualize how the generator distribution \( p_g \), the discriminator output \( D(x) \), as well as the true data distribution \( p_{\text{data}} \) evolve over the course of training. After random initialization, the generator produces unrealistic samples and its distribution \( p_g \) is clearly distinct from the true data distribution \( p_{\text{data}} \). At this stage, the discriminator is not yet trained and its output \( D(x) \) is inconsistent. Shortly after training begins, the discriminator learns to reliably distinguish real from fake samples and \( D(x) \) becomes more stable. As training progresses, the generator improves and its output distribution \( p_g \) gradually aligns with \( p_{\text{data}} \). At convergence, \( G \) successfully matches the real data distribution, and the output of \( D \) approaches 0.5 for all inputs, indicating that it can no longer distinguish real from generated samples (Figure~\ref{fig:training_distributions}).

\subsubsection{Results and Discussion}

The original GAN paper evaluated the model using both qualitative visual inspection and quantitative log-likelihood estimates. Since the GAN framework lacks an explicit data distribution \( p(x) \), likelihood-based evaluation requires fitting a Gaussian kernel to generated samples. Although this method is noisy and unreliable in high dimensions, it remains one of the few options for evaluation. On the MNIST dataset, GANs achieve a Parzen window-based log-likelihood of \( 225 \pm 2 \). On the Toronto Face Dataset (TFD), GANs also perform competitively (in 2014) with a log-likelihood of \( 2057 \pm 26 \). These metrics indicate that GANs are capable of capturing complex data distributions effectively. Visual samples from MNIST, TFD and CIFAR-10 confirm that the model generates diverse and plausible samples.

Vanilla GANs offer several advantages. They eliminate the need for inference or MCMC sampling, relying entirely on backpropagation. This makes them computationally efficient while producing sharp, high-quality outputs. Furthermore, GAN allows for architectural flexibility, as both \( G \) and \( D \) can be implemented as arbitrary, differentiable functions.

However, training GANs remains challenging. The adversarial process is prone to instability and can suffer from \textit{mode collapse}, where the generator maps many different latent vectors \( z \) to the same output \( x \), resulting in limited diversity. Another issue occurs when \( D \) becomes too strong early on and rejects generated samples with high confidence. This causes \( G \) to receive vanishing gradients, preventing effective learning. The absence of an explicit likelihood complicates evaluation, while successful training depends on carefully balancing the learning dynamics of both networks.

\subsubsection{Evolution of the proposed architecture}

Important to note is that the vanilla GAN architecture is from 2014 and is one of the earlier generative models. Since then many follow-up works have been published, achieving better training stability, higher image resolution and diversity utilizing bigger more intricate architectures. In the following paragraphs, four of these works are analyzed in regard to how GAN problems have been solved.

\paragraph{Progressive Growing of GANs~\cite{karras2018progressivegrowinggansimproved}} is a key architectural innovation introduced to improve training stability and generate high-resolution images. Instead of training at full resolution from the start, ProGAN begins by generating low-resolution images (\( 4 \times 4 \) px) and incrementally increases the resolution by adding layers to both generator and discriminator. This progressive approach has two main advantages: It stabilizes training by starting with a simplified task and it reduces computational cost. As training progresses and new layers are added, earlier layers are fine-tuned and the network gradually adapts to higher resolutions. To avoid destabilization of the training when switching to a new resolution, ProGAN uses a fade-in mechanism where new layers are linearly blended with existing layers over several training steps using a parameter \( \alpha \in [0,1] \). ProGAN also introduces further improvements: equalized learning rates and pixel-wise feature normalization contribute to more stable convergence. The authors evaluated ProGAN on high-resolution data sets such as CelebA-HQ (up to \( 1024 \times 1024 \) px) and LSUN bedrooms \( 256 \times 256 \) px), reporting unprecedented image fidelity at the time. Generated images are evaluated using the Fréchet Inception Distance (FID), which compares the distribution of generated features to those of real images using an inception network. Lower FID scores indicate higher quality and diversity of outputs~\cite{heusel2018ganstrainedtimescaleupdate}. ProGAN achieves significantly better FID than previous GAN variants, especially at high resolutions. In general, ProGAN represents a shift towards scalable and stable GAN training for high-resolution image generation.

\paragraph{Progressive Augmentation GAN~\cite{zhang2019progressiveaugmentationgans}} addresses the problem of discriminator overfitting, which can lead to vanishing gradients for the generator. Instead of directly manipulating image data, PA-GANs progressively increase the difficulty of \( D \)s task by adding random binary bits to real and generated inputs.

Training begins as a vanilla GAN. After a fixed number of epochs, one augmentation bit is added to \( D \)'s input, increasing the task complexity and preventing the discriminator from becoming too powerful too early. This ensures that \( G \) continues to receive informative gradients throughout training. PA-GAN integrates seamlessly with existing GAN architectures and shows improved FID scores~\cite{heusel2018ganstrainedtimescaleupdate} as well as more stable training with minimal computational overhead. PA-GANs offer a simple yet effective regularization strategy. However, they still face mundane GAN limitations such as \textit{mode collapse} and hyperparameter sensitivity, especially in generator learning.

\paragraph{Energy-based GANs (EB-GAN)~\cite{zhao2017energybasedgenerativeadversarialnetwork}} reinterpret the discriminator as an energy function rather than a probabilistic classifier. Here, \( D \) is an autoencoder that assigns low energy (L2 reconstruction error - MSE) to real samples and high energy to generated ones. This shifts the training dynamic to energy minimization for real samples and energy maximization for fake samples, improving training stability and gradient feedback. The generator is trained to produce samples with low reconstruction loss, while a margin-based loss function ensures that only samples below a threshold \( m \) affect \( D \)'s updates. The energy landscape also implicitly defines a Boltzmann-like distribution:

\[
p(x) = \frac{1}{Z} \exp(-E(x)), \quad \text{where} \quad Z = \int \exp(-E(x)) \, dx
\]

Here, \( p(x) \) defines the probability density assigned to a sample \( x \) based on its energy \( E(x) \). \( Z \) is the partition function that normalizes the distribution. However, computing \( Z \) requires integrating over the entire input space, which is computationally infeasible in high dimensions. Nevertheless, this architectural change improves training stability and avoids saturation issues. It performs well on datasets such as MNIST and CelebA, but remains sensitive to the margin parameter \( m \) and may still suffer from limited sample diversity.

\paragraph{BigGAN~\cite{brock2019largescalegantraining}} marks a major milestone in generative modeling by showing that scaling with four times more model parameters and eight times batch size significantly improves performance. It extends standard GANs not only with deeper and wider architectures but also with class-conditional inputs, as well as other optimizations for large-scale training.

Both \( G \) and \( D \) are class-conditional, using label embeddings merged with features via conditional batch normalization. Built on deep residual networks (ResNets~\cite{he2016deep}), BigGAN benefits from improved gradient flow. The model also integrates orthogonal regularization, shared embeddings and a truncation trick to balance image quality and diversity. The latent vector \( z \) is sampled from a truncated normal distribution, typically restricted to the standard deviation \( \pm 1 \). This focuses on sampling high-density regions of the latent space, enhancing image fidelity and sharpness at the cost of reducing output diversity. The truncation threshold can be tuned to balance out these effects.

A core innovation is the use of large batch sizes (up to 2048) and spectral normalization in both networks, enabling stable training. BigGAN is trained on ImageNet at resolutions of up to \( 512 \times 512 \) px and achieves state-of-the-art FID scores~\cite{heusel2018ganstrainedtimescaleupdate}. Although BigGAN delivers unmatched realism, its high computational cost and sensitivity to hyperparameters pose accessibility challenges for smaller research groups.

\subsubsection{Conclusion and Comparison to Transformers}

Generative Adversarial Networks have reshaped generative modeling by enabling sharp, high-resolution image synthesis through adversarial feedback. Architectural innovations like ProGAN and BigGAN have addressed major challenges such as training instability, \textit{mode collapse} and limited resolution, making GANs powerful but sensitive to hyperparameter tuning and difficult to evaluate due to their implicit likelihood.

Transformer-based models have emerged as a strong alternative, using self-attention and large-scale pre-training to model complex data distributions. Unlike GANs, they optimize log-likelihood directly, allowing stable training and structured output generation across text, image and audio modalities. However, they require significantly more compute and memory and inference is slower compared to GANs’ holistic, single-forward-pass generation~\cite{dhiwise2025gantransformer}.

GANs remain ideal for fast, domain-specific generation on limited training data, while Transformers offer better control, scalability and multi-modal capabilities. Promising future directions include hybrid models combining both paradigms, as well as advances in quantum GANs, 3D scene synthesis, biomedical privacy-preserving GANs and harder-to-engineer multimodal systems.

\newpage

\subsection{Latent Diffusion Models} %RADU

The following subsection will focus on explaining image generation using Latent Diffusion Models ~\cite{rombach2022high}, a technology based on Diffusion Models ~\cite{ho2020denoising}, aiming to deliver improvements in terms of training and inference efficiency compared to previous image generation methods.

\subsubsection{Background and Motivation}

\begin{figure}[h]
    \centering
    \includegraphics[width=0.70\textwidth]{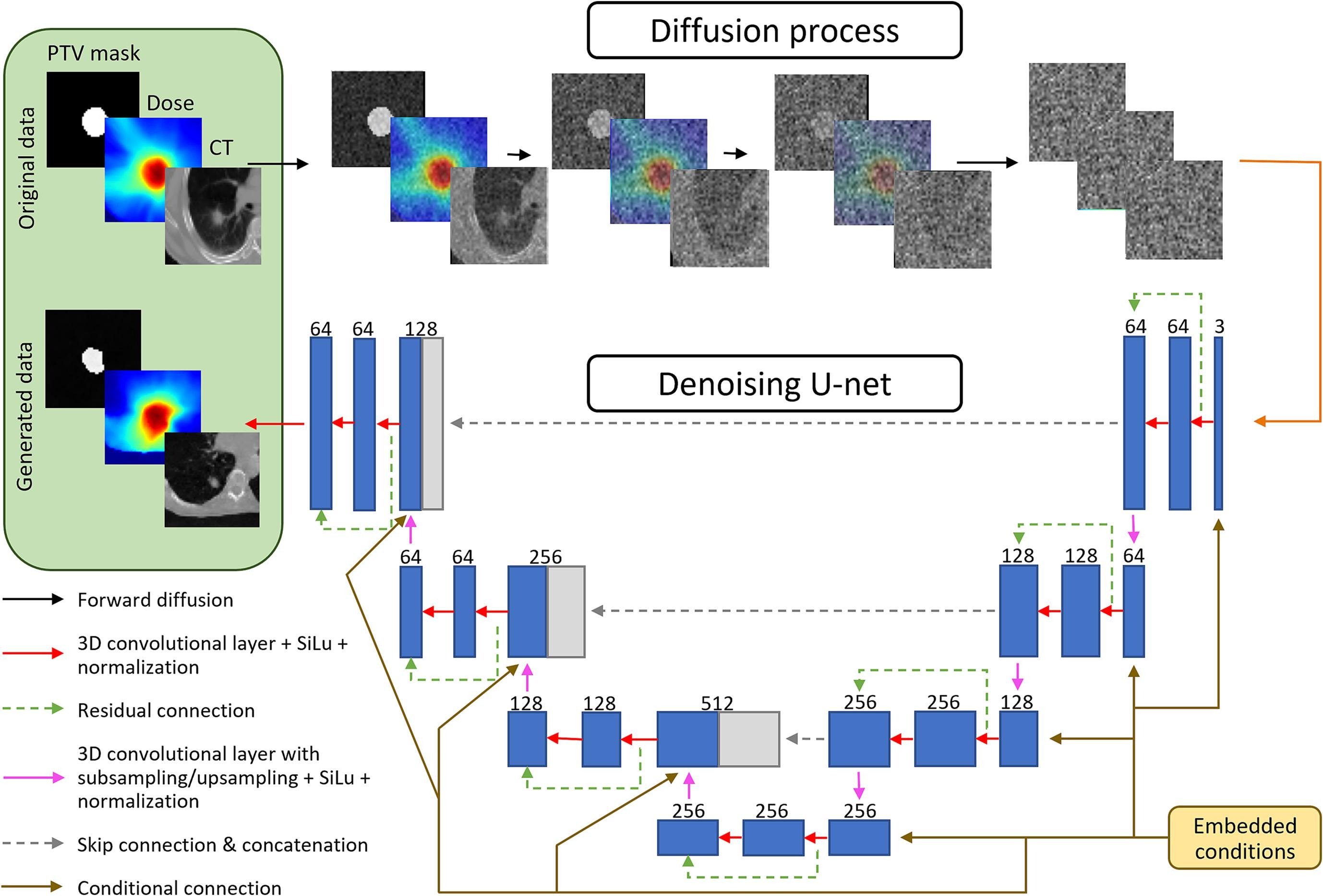}
    \caption{
    Diffusion Model architecture. 
    \label{fig:DMs}
    }
\end{figure}

As image generation becomes an increasingly prominent computer vision task, the computational expense of training and inference has emerged as a critical bottleneck for neural networks developed to handle this task. Although well-known neural networks for image generation, such as Generative Adversarial Networks (GANs) ~\cite{goodfellow2014generative} and Diffusion Models (DMs) ~\cite{ho2020denoising}, offer promising results, they struggle with computational inefficiency. To address this, Latent Diffusion Models (LDMs) ~\cite{rombach2022high} were introduced in 2022. LDMs are a type of neural network designed to tackle the lengthy training and inference times of previous models.
\\
\\
Latent Diffusion Models are closely related to Diffusion Models. Therefore, understanding how DMs work is crucial for understanding the concepts behind LDMs. DMs generate images by taking a random noise distribution and gradually removing the noise until a clear image that matches the desired output, given a conditioning prompt, is produced. This can be thought of as the inverse process of gradually adding noise to an image. DM training involves two key steps: the diffusion process and the denoising process, as we can see in Figure \ref{fig:DMs}. In the diffusion process, Gaussian noise is incrementally added to the original image using Markov Chain transitions. Following this, in the denoising process, a U-Net learns to identify and remove the added noise, aiming to reconstruct the original image. This is a self-supervised learning process because the ground truth is the original image itself. Although DMs produce high-quality results, they are computationally very expensive because they operate directly in the high-dimensional pixel space, leading to long inference times. In particular, generating 10,000 samples can take up to a day with a conventional DM.

\subsubsection{Approach and Methodology}

Latent Diffusion Models promise to solve this problem by moving the training and image generation from the high-dimensional pixel space into a lower-dimensional latent space. This is achieved through two main stages: perceptual compression and semantic compression. During perceptual compression, an autoencoder is trained on a large dataset of images. This step is vital because it establishes a perceptual equivalence between the pixel space and the latent space of the autoencoders. This latent space is a core component of LDMs, as it allows the computationally intensive operations to be performed in a more manageable lower-dimensional space. Once the autoencoder is trained, we have a trained encoder and decoder.
\\
\\
The next stage is semantic compression. Here, the Diffusion Model described earlier is applied within the latent space. By training the DM in this smaller, latent space instead of the high-dimensional pixel space, LDMs achieve significantly improved efficiency compared to conventional Diffusion Models.

\begin{figure}[h]
    \centering
    \includegraphics[width=0.85\textwidth]{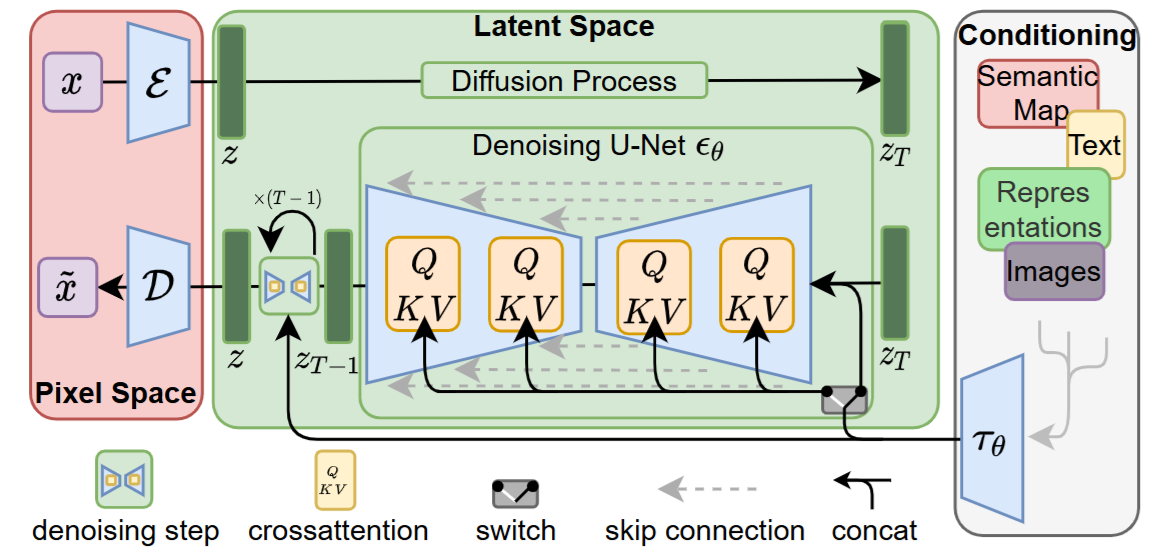}
    \caption{
    Latent Diffusion Model architecture. In courtesy of \cite{rombach2022high}.
    \label{fig:LDMs}
    }
\end{figure}

The architecture of a Latent Diffusion Model is illustrated in Figure \ref{fig:LDMs}. The pixel space operations, handled by the encoder and decoder, are shown in red. The architecture within the latent space mirrors that of a conventional Diffusion Model, in green. It is important to note that the network is trained on a text-image database consisting of text-image pairs. After the initial training of the autoencoder on the images, the diffusion process is performed for each image within the latent space, adding noise. Subsequently, the Denoising U-Net is trained in this latent space. It attempts to reconstruct a less noisy version of the image at each of the T time steps of the Markov Chain diffusion process. Crucially, during the U-Net's training, a domain-specific encoder translates the text labels from the dataset into the latent space. This text representation is then concatenated with the noisy image representation in the latent space and also injected into the U-Net's layers via an attention mechanism. This technique trains the Denoising U-Net in a prompt-conditioned manner, enabling it to generate images based on a text input.
\\
\\
For the inference stage, only three components are needed: the domain-specific encoder, the Denoising U-Net, and the Decoder. To generate an image, the input prompt is first encoded into the latent space. Then, starting with a concatenation of the encoded prompt and a random noise image in the latent space, the U-Net iteratively removes the noise to generate a final image representation in the latent space. This representation is then decoded back into the pixel space by the Decoder.
\\
\\
In summary, Latent Diffusion Models are essentially Diffusion Models applied within the latent space of a pre-trained autoencoder. This approach yields a significant boost in efficiency, as the model requires far less memory in both training and inference, making them faster. Another advantage of LDMs is the re-usability of the autoencoder. Due to the modular training process (perceptual compression followed by semantic compression), the autoencoder is independent of the Denoising U-Net and can be repurposed for other tasks.

\subsubsection{Experiments and Results}

To test the efficiency of LDMs, different down-sampling factors \textit{f} in the order of \textit{ \{1, 2, 4, 8, 16, 32\}} for the perceptual compression were tested, annotated as \textit{LDM-f}. To obtain comparable test results, the computational resources were fitted to a single NVIDIA A100 GPU for all experiments. An obvious result was the fact that small downsampling factors for \textit{LDM-{1,2}} result in a slow training process, while large values of \textit{f} cause stagnation in fidelity after comparatively few training steps.
\\
\\
An important aspect under investigation during the experiments was the comparison with DMs. Training powerful DMs in pixel space typically consumes hundreds of GPU days (e.g., 150-1000 V100 days). Generating 50,000 samples can take approximately 5 days on a single A100 GPU in pixel space. LDMs significantly reduce these computational costs. For instance, in inpainting, LDMs showed a speed-up of at least 2.7x compared to pixel-based DMs while also improving FID scores by at least 1.6x (a metric used to evaluate the quality of images generated).
\\
\\
LDMs also achieved new state-of-the-art FID scores on the CelebA-HQ dataset, outperforming previous likelihood-based models as well as GANs.
They also demonstrated highly competitive performance on other datasets like FFHQ, LSUN-Churches, and LSUN-Bedrooms, outperforming prior diffusion-based approaches on most datasets. For example, on LSUN-Bedrooms, LDMs achieved a score close to DM while using half its parameters and 4-times less training resources.
\\
\\
Regarding text-to-image synthesis, a 1.45B parameter LDM trained on the LAION-400M dataset demonstrated powerful generalization to complex, user-defined text prompts. On the MS-COCO dataset, the guided LDM achieved performances on par with recent state-of-the-art autoregressive and diffusion models (e.g., Make-A-Scene, GLIDE), despite substantially reducing parameter count.

\subsubsection{Discussion and Conclusion}

Although LDMs offer better computational performance than conventional DMs, there still are limitations and potential further improvements to training and inference times. Furthermore, the generated images can sometimes lack high-frequency details, such as accurately rendering fingers on a human hand. An additional concern is the unknown extent to which these models memorize or reveal training data, raising questions about data privacy. On an ethical level, the ability of LDMs to generate highly realistic deepfakes poses a significant problem, as it can fuel disinformation campaigns.
\\
\\
In conclusion, LDMs represent an important milestone in the field of image generation. Their key innovation is applying conventional DMs in a latent space rather than the pixel space, which has had a major impact on reducing computation time. However, there is still room for improvement in image quality, and significant ethical questions surrounding this technology remain open.

\newpage
\section{Self-Supervised Representation Learning }

Self-Supervised Learning (SSL) has emerged as a crucial paradigm in Machine Learning, bridging the gap between supervised and unsupervised approaches. Instead of relying on vast amounts of manually labeled data as in supervised learning, SSL leverages the inherent structure within unlabeled data to create its own supervisory signal. This is achieved by defining a "pretext task", where the model learns to predict certain properties of the data itself, such as a missing part or a transformation that has been applied. By solving these pretext tasks, the model learns meaningful and robust feature representations that can then be fine-tuned for various downstream tasks like image classification or object detection with much less labeled data.
\\

This ability to learn from abundant unlabeled data is particularly significant for training large-scale models like Vision Transformers, which, as discussed earlier, are data-hungry~\cite{dosovitskiy2020image}. Self-supervised learning helps to instill in these models the kind of inductive biases that are naturally present in architectures like CNNs, improving their performance and scalability.
\\

Self‑supervised visual representation learning can be broadly grouped into three families: contrastive, predictive/distillation, and generative/reconstructive approaches. Contrastive methods, exemplified by SimCLR~\cite{chen2020simpleframeworkcontrastivelearning}, learn by pulling augmented views of the same image together while pushing representations of different images apart. Predictive or distillation‑based techniques such as DINO~\cite{caron2021emerging} train a student network to match the outputs of a momentum‑updated teacher across multiple crops, eliminating the need for explicit negative pairs and therefore lying outside the contrastive category. Generative methods such as Masked Autoencoders (MAE)~\cite{he2022masked} task the model with reconstructing an original input from a corrupted version. This forces the model to learn the underlying data distribution to fill in the missing information.
\\
\\
The following sections will cover DINO and MAE as the primary representatives of predictive distillation and reconstructive strategies, respectively.

\subsection{DINO: Self-Distillation with No Labels}
% \subsection{Emerging Properties in Self-Supervised Vision Transformers~\cite{caron2021emerging}}

The following section discusses DINO~\cite{caron2021emerging} (self-\textbf{DI}stillation with \textbf{NO} labels), a seminal framework for non-contrastive representation learning.

\subsubsection{Background and Motivation}

Transformers~\cite{vaswani_2017_attention} have emerged as a highly influential building block in NLP, leading to a number of breakthroughs. Naturally, there has been a lot of interest in adapting transformers to the field of Computer Vision (CV), with the intent of achieving similar breakthroughs as in Natural Language Processing (NLP). A front-runner model in this space is the Vision Transformer~\cite{dosovitskiy2020image}, or ViT.
\\
\\
When first introduced, the ViT achieved state-of-the-art performance but required a vast amount of data to surpass traditional CNN-based models. It was hypothesized that ViTs only excelled when trained on large datasets because the additional scale allowed them to learn inductive biases, which were already present in CNN-based models.
\\
\\
This challenge can be addressed by borrowing techniques from NLP. Self-supervised learning, particularly by means of language modeling, has proven to be highly effective in constructing foundation models (e.g., GPT, BERT ~\cite{devlin-etal-2019-bert}). When transferred to CV, self-supervision can help bring CNNs' inductive biases to transformers without the need for large-scale labeled datasets.
\\
\\
In essence, DINO is a framework for representation learning. A key difference between it and common representation learning techniques such as MoCo~\cite{he2020momentum}, SimCLR~\cite{chen2020simpleframeworkcontrastivelearning}, is that DINO employs non-contrastive learning. Models trained under DINO learn rich, descriptive embeddings, useful for downstream tasks. It is compatible with any kind of model, whether convolutional, transformer, or otherwise.

\subsubsection{Approach and Methodology}

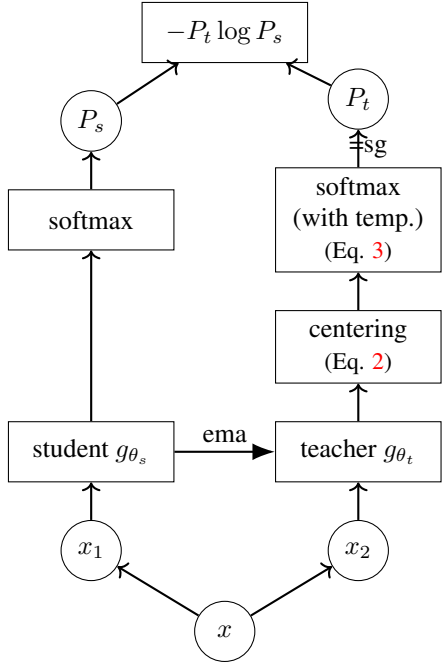
\begin{figure}
    \centering
    \begin{tikzpicture}[
      node distance=0.5cm and 1.2cm,
      every node/.style={font=\normalsize},
      box/.style={draw, minimum width=2.2cm, minimum height=0.8cm, align=center},
      round/.style={draw, circle, minimum size=0.8cm},
      arr/.style={->, thick},
      thickarrow/.style={-{Latex[length=3mm]}, thick}
    ]
    
    % Input x
    \node[round] (x) {$x$};
    
    % Augmented views
    \node[round, above left=of x] (x1) {$x_1$};
    \node[round, above right=of x] (x2) {$x_2$};
    
    % Teacher branch (right side)
    \node[box, above=of x2] (teacher) {teacher $g_{\theta_t}$};
    \node[box, above=of teacher] (center) {centering\\{\small (Eq. \ref{eq:dino-centering})}};
    \node[box, above=of center] (softmax2) {softmax\\(with temp.)\\{\small (Eq. \ref{eq:dino-softmax})}};
    \node[round, above=of softmax2] (p2) {$P_t$};
    
    % Student branch (left side)
    \node[box, above=of x1] (student) {student $g_{\theta_s}$};
    \node[box] at (softmax2 -| student) (softmax1) {softmax};
    \node[round, above=of softmax1] (p1) {$P_s$};
    
    \path (p2) -- (p1) node[pos=0.5, midway] (secret) {};
    
    \node[box, above=of p1, above=of secret] (loss) {$-P_t\log P_s$};
    
    % Arrows
    \draw[arr] (x) -- (x1);
    \draw[arr] (x) -- (x2);
    
    \draw[arr] (x1) -- (student);
    \draw[arr] (student) -- (softmax1);
    \draw[arr] (softmax1) -- (p1);
    
    \draw[arr] (x2) -- (teacher);
    \draw[arr] (teacher) -- (center);
    \draw[arr] (center) -- (softmax2);
    \draw[arr] (softmax2) -- (p2) node[pos=0.5, midway]{\ \ \ =sg};
    
    \draw[arr] (p2) -- (loss);
    \draw[arr] (p1) -- (loss);
    
    % EMA arrow
    \draw[thickarrow] (student.east) -- node[above] {ema} (teacher.west);
    
    \end{tikzpicture}
    \caption{DINO illustration}
    \label{fig:dino-illustration}
\end{figure}

Figure~\ref{fig:dino-illustration} concisely describes the framework. For the purpose of describing the approach, assume that the training goal is to learn embeddings for images. From each training sample $x$, a set $x_1$ of local views (small crops) is extracted, and a set $x_2$ of local and global views (large crops) is extracted.
\\
\\
The set $x_1$ is passed through the student model, followed by a softmax operation, yielding the student's probability distribution $P_s$. The set $x_2$ is passed through the teacher model, followed by a centering operation (Equation \ref{eq:dino-centering}), followed by a softmax with temperature for sharpening (Equation \ref{eq:dino-softmax}), yielding the teacher's probability distribution $P_t$ (Equation \ref{eq:dino-teacher-pt}). The two distributions are used to compute cross-entropy loss $H(P_t,P_s)=-P_t\log P_s$, which encourages the student's probabilities to align with the teacher's.
\begin{equation}
P_t=\operatorname{softmax}(\tau;g_{\theta_t}(x_2)-c)
\label{eq:dino-teacher-pt}
\end{equation}
\\
The mean for the centering operation is computed by tracking the per-feature exponential moving average $c$ of the teacher's outputs.
\begin{equation}
c\leftarrow mc+(1-m)\frac{1}{B}\sum_{i=1}^B g_{\theta_t}(x_i)
\label{eq:dino-centering}
\end{equation}
\\
Softmax with temperature is computed by dividing all dimensions of the input vector $x$ with a temperature $\tau$. Temperature controls the sharpness of the resulting probability distribution. A high temperature produces a softer (more uniform) distribution, while a low temperature yields a sharper (more peaked or degenerate) one.
\begin{equation}
\operatorname{softmax}_i(\tau;x)=\frac{\exp{(x_i/\tau)}}{\sum_j\exp{(x_j/\tau)}}
\label{eq:dino-softmax}
\end{equation}
\\
The teacher's parameters $\theta_t$ are computed by tracking the exponential moving average of the student's parameters $\theta_s$ (Equation \ref{eq:dino-teacher}). The stop-gradient (or sg) operator on the teacher's half of the network prevents gradient updates to the teacher's parameters.
\begin{equation}
\theta_t\leftarrow\lambda\theta_t+(1-\lambda)\theta_s
\label{eq:dino-teacher}
\end{equation}
\\
The student's parameters $\theta_s$ are computed by way of stochastic gradient descent.

\subsubsection{Experiments}

In this section, the accuracy of models was measured by computing embeddings for images in the validation set and fitting a k-NN classifier on those embeddings.
\begin{figure}
    \centering
    \includegraphics[width=0.2\linewidth]{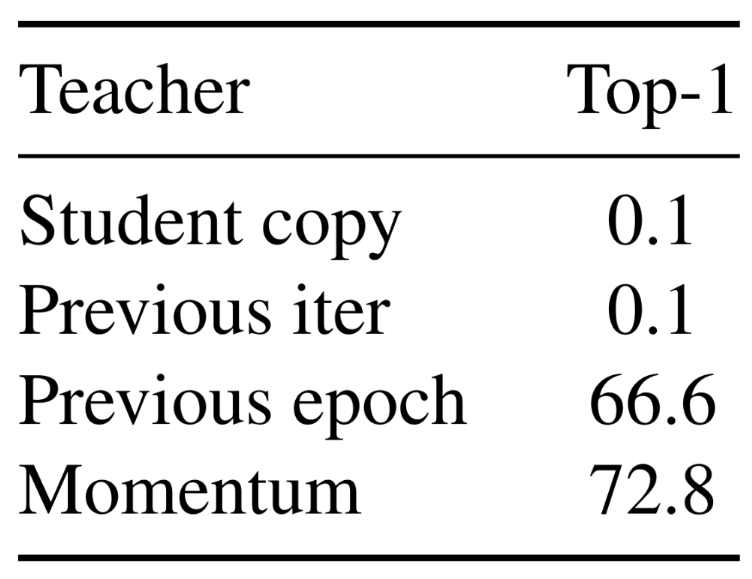}
    \caption{Comparison of accuracy achieved with differently-constructed teachers.}
    \label{fig:dino-teachers}
\end{figure}
\\
\\
\textbf{Building different teachers from the student.} A series of different strategies for determining the teacher's parameters from the student's were evaluated. The results are listed in Figure~\ref{fig:dino-teachers}. Of the strategies that were tried, the momentum teacher performed the best.
\begin{figure}
    \centering
    \includegraphics[width=0.25\linewidth]{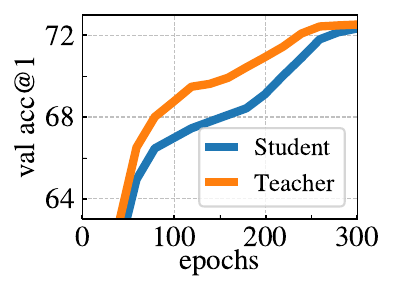}
    \caption{Accuracies of student and teacher throughout training.}
    \label{fig:dino-student-vs-teacher}
\end{figure}
\\
\\
\textbf{Examining the momentum teacher.} Measuring the accuracy achieved by the student and momentum teacher throughout training (Figure~\ref{fig:dino-student-vs-teacher}) reveals that the momentum teacher tends to outperform the student. A possible explanation for this phenomenon is that the momentum teacher acts as an ensemble of models. Calculating the EMA of the student at every training iteration can be interpreted as a form of repeated Polyak-Ruppert averaging. This explanation supports the interpretation of DINO as a form of self-distillation.
\begin{figure}
    \centering
    \includegraphics[width=0.45\linewidth]{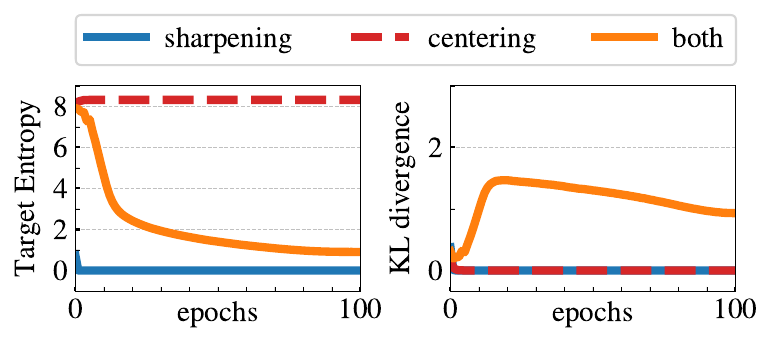}    \captionsetup{width=0.6\linewidth}
    \caption{Left: evolution of entropy of teacher's output; Right: evolution of KL-divergence of student's output from teacher's}
    \label{fig:dino-entropy-kl-divergence}
\end{figure}
\\
\\
\textbf{Studying the effect of sharpening and centering operations.} To better understand the effect of sharpening and centering on training, three configurations are studied: using only sharpening, only centering, or both. To interpret the training dynamic under the three configurations more effectively, it is useful to decompose the cross-entropy loss into a sum of the entropy of $P_t$ and the KL-divergence of $P_s$ from $P_t$:
\begin{equation}
H(P_t,P_s)=h(P_t)+D_{KL}(P_t\|P_s)
\end{equation}
\\
Results of training under each configuration are presented in Figure~\ref{fig:dino-entropy-kl-divergence}. Using only sharpening, entropy becomes very low, indicating a degenerate distribution; using only centering, entropy becomes very high, indicating a uniform distribution. Moreover, the KL-divergence in both cases is close to zero, indicating that the student and teacher collapse to producing essentially the same distribution.
\\
\\
When both operations are used together, entropy and KL-divergence are maintained within an appropriate range. This suggests that the operations can be interpreted as a form of entropy control.

\subsubsection{Results and Discussion}

\begin{figure}
    \centering
    \includegraphics[width=0.5\linewidth]{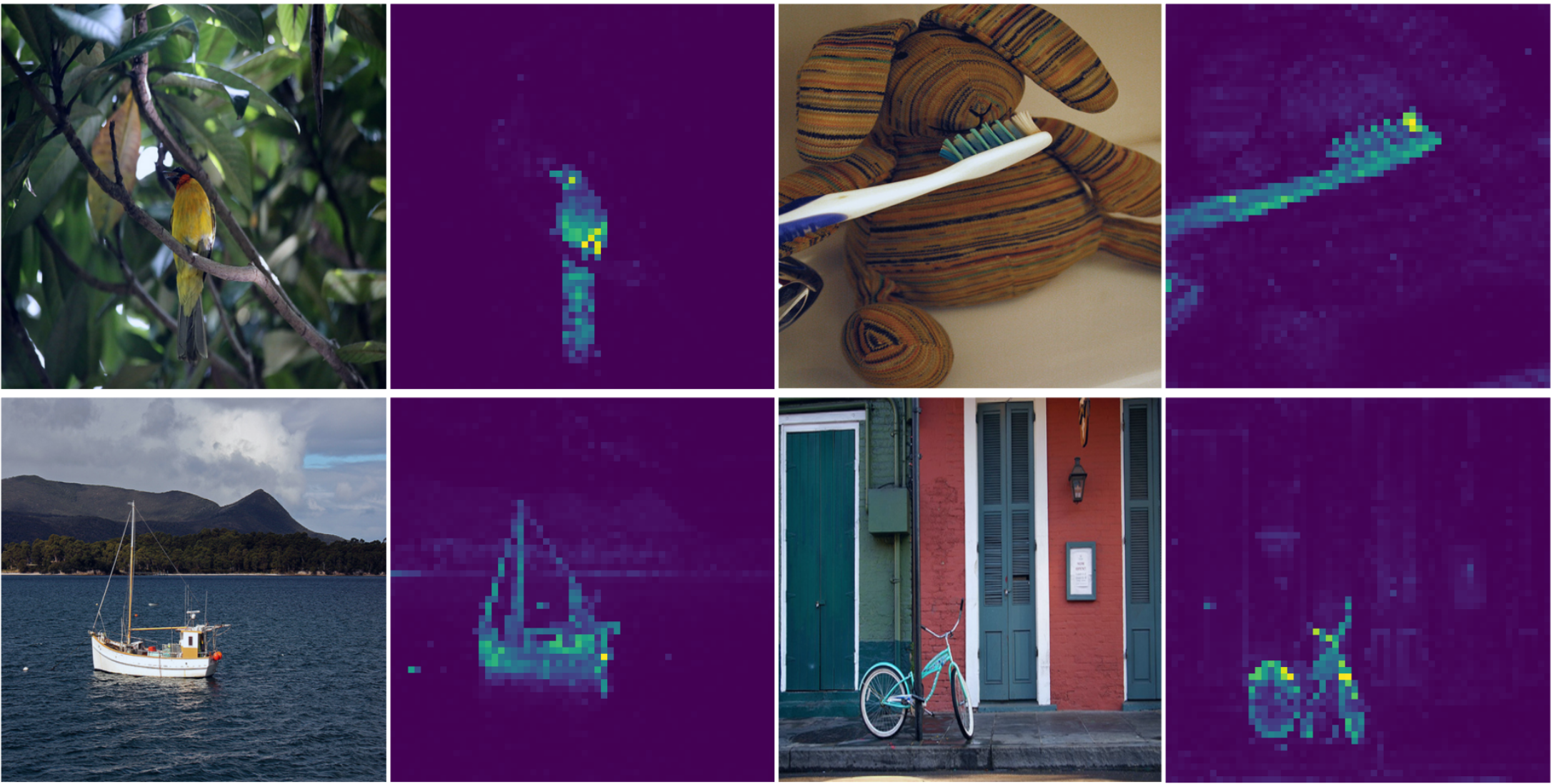}
    \caption{Self-attention from a Vision Transformer trained under DINO. In courtesy of the authors in~\cite{caron2021emerging}.}
    \label{fig:dino-self-attention}
\end{figure}
Models trained under the DINO framework develop rich, representative embeddings. At the time of release, performance was state-of-the-art, achieving up to 80.1\% top-1 accuracy on ImageNet, using a ViT-B/8 with a final fine-tuned linear layer. Inspecting the self-attention patterns of ViTs trained under DINO yields especially surprising results. Despite having been trained with no supervision, the self-attention maps look remarkably like segmentation maps (Figure~\ref{fig:dino-self-attention}).
\\
\\
In conclusion, DINO is a seminal work in non-contrastive representation learning for computer vision. It took an important step towards reducing the quantity of labeled data necessary for training high-performing ViTs. It popularized self-distillation and entropy control techniques, paving the way for future models such as iBOT~\cite{zhou2022ibotimagebertpretraining}, EsViT~\cite{li2022efficientselfsupervisedvisiontransformers}, and its direct successor DINOv2~\cite{oquab2024dinov2learningrobustvisual}.

\newpage
\subsection{Masked Autoencoders}
%\subsection{\textit{Masked Autoencoders Are Scalable Vision Learners~\cite{he2022masked}}} %NEIL
\begin{figure}[!b]
    \centering
    \includegraphics[width=0.75\linewidth]{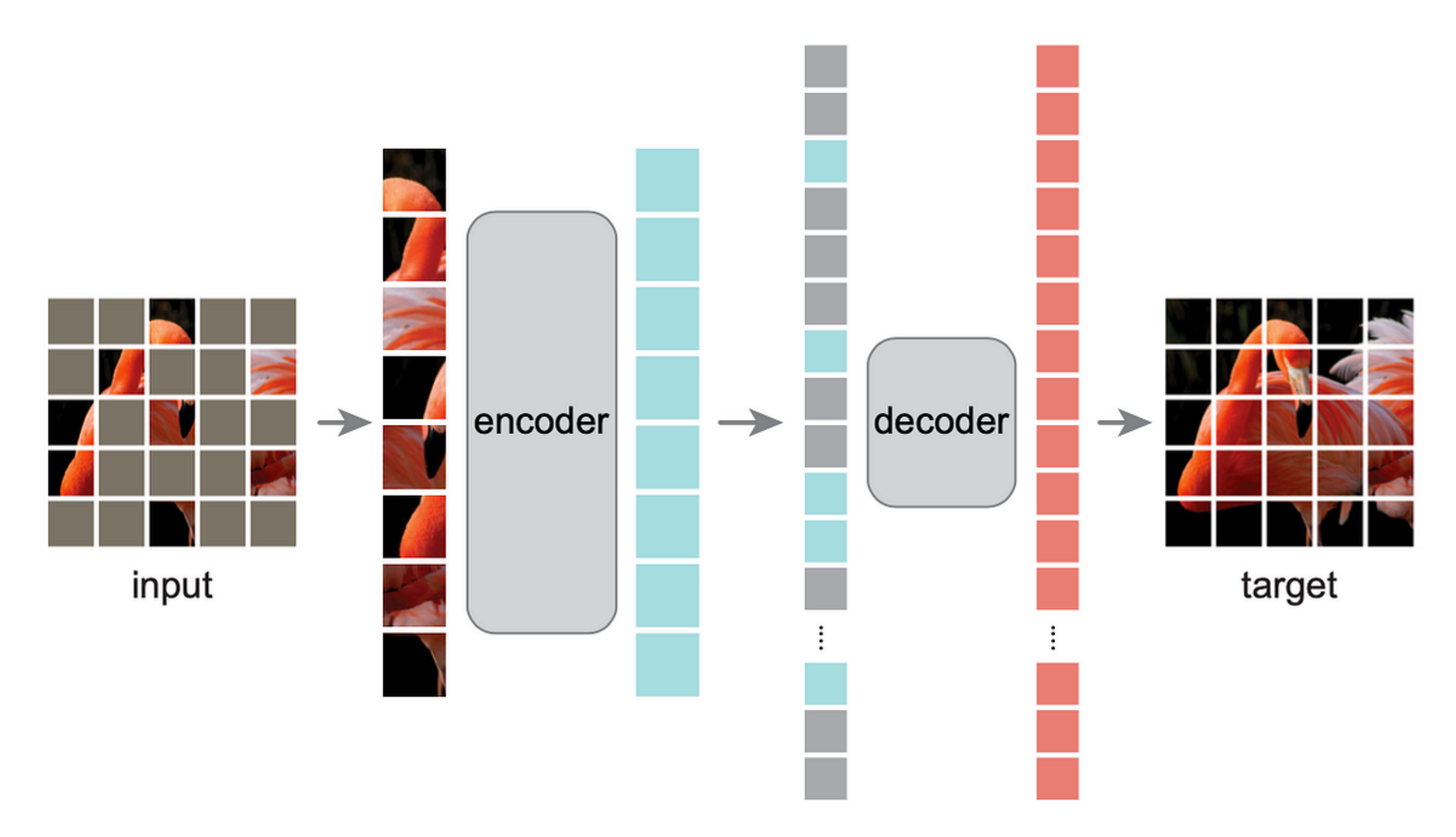}
    \caption{MAE architecture. In courtesy of \cite{he2022masked}.}
    \label{fig:mae_arch}
\end{figure}

This last section covers Masked Autoencoders, a paradigm shift in self supervised learning that lead to SOTA results in the most efficient and scalable manner to date.
\subsubsection{Background and Motivation}

Self-supervised representation learning has driven major advances in natural language processing and is increasingly important in computer vision. In NLP, masked token prediction (as popularized by BERT \cite{devlin-etal-2019-bert}) enabled training extremely large transformers on unlabeled text by hiding some tokens and requiring the model to infer them from context. Adapting this idea to vision is promising but non-trivial due to key differences between images and text. Images are highly redundant: much of a scene’s content can be guessed from neighboring pixels. Early vision self-supervision focused on discriminative or contrastive objectives (instance classification or feature invariances) to avoid trivial pixel-level reconstruction. For example, methods like MoCo \cite{he2020momentum} and DINO \cite{caron2021emerging} learn by making representations of different augmented views of an image similar, rather than by generating pixels. These approaches yielded good features but involve complex training schemes (e.g. momentum encoders, large batches, or teacher networks) and may saturate with larger models or longer training. Vision Transformers (ViT) \cite{dosovitskiy2020image} recently opened the door to more flexible token-based image modeling, similar to text. He \textit{et al.} \cite{he2022masked} propose Masked Autoencoders (MAE) to leverage this opportunity. The motivation is to design a simple yet scalable self-supervised vision learner, analogous to BERT for images, that can learn massive transformer models without labels. Two main insights form the basis of MAE’s design: (1) use an extremely high masking ratio on image patches to create a challenging pretext task that forces the model to learn meaningful global structure (addressing vision’s high redundancy), and (2) use an asymmetric encoder–decoder architecture to efficiently handle the large mask percentages. This approach promises to learn highly capable models that generalize well, bridging the gap between vision and language self-supervision in terms of scalability and performance.

\subsubsection{Approach and Methodology}

MAE is a Vision Transformer-based autoencoder that learns by reconstructing missing parts of the input image. The overall architecture is illustrated in Figure~\ref{fig:mae_arch}. The method can be summarized in four key steps:

\begin{enumerate}\item \textbf{Random high-rate masking:} Each input image is divided into fixed-size patches (e.g. $16\times16$ pixels as in ViT). A large random subset of these patches (typically 75\%) is masked out (removed), leaving only 25\% visible patches. Such an aggressive masking ratio creates a difficult pretext task, drastically reducing low-level redundancy in the input.
\item \textbf{Encoding visible patches:} The subset of visible patches is passed through an encoder (a ViT) that operates only on these unmasked patches. Importantly, no placeholder tokens are used for the masked positions at this stage – the encoder sees a partial image. Positional embeddings are still applied so that the encoder’s output is aware of patch locations. By processing only 25\% of tokens, the encoder’s computation is greatly reduced, which makes training far more efficient than if the full image token sequence was used.
\item \textbf{Inserting mask tokens:} After encoding, the model reinserts representations for the missing patches in the form of learned mask tokens. These mask tokens (one per missing patch position) are appended to the encoder’s output sequence. This restores the sequence to its original length (all patches), where visible patch embeddings are interleaved with mask token placeholders at their proper original positions.
\item \textbf{Decoding and reconstruction:} A lightweight decoder transformer then processes the full sequence (visible patch embeddings + mask tokens) and attempts to reconstruct the original image in the pixel space. The decoder predicts pixel intensities for each patch. A reconstruction loss (mean squared error) is computed only on the masked patches (the model is not penalized for re-predicting the pixels it already saw). After pre-training, the decoder is discarded; only the encoder is used for downstream tasks by prepending it with a task-specific head or by fine-tuning it on full images.
\end{enumerate}

This asymmetric design, which contains a heavy full-capacity encoder applied to just the visible patches, together with a smaller decoder handling all tokens, is a core feature of MAE. By shifting the computational load of processing mask tokens to the decoder, the encoder avoids seeing a majority of tokens, yielding a significant speed-up (the authors report a $3\times$ reduction in pre-training time, for the same number of epochs. In practice, the encoder can be a big ViT (e.g., ViT-Large or Huge) while the decoder is much narrower and shallower (e.g., 8 Transformer blocks vs. 24 in the encoder. The high masking ratio is really important: Authors show that masking about 75\% of patches hits a sweet spot, making the task challenging enough to prevent simply filling in details from low-level cues, yet still learnable. This is in contrast to BERT’s typical 15\% masking; vision required a much higher fraction of missing content to yield semantic learning. In particular, masks are sampled randomly from each image; experiments found that this simple random masking outperforms structured masking strategies, presumably because it avoids predictable gaps and covers diverse regions. MAE’s reconstruction target is the raw pixel values (usually normalized) of the masked patches. This choice keeps the approach simple. Unlike some earlier methods (e.g., BEiT \cite{bao2021beit}) that use a pre-trained dVAE tokenizer to predict discrete visual tokens, MAE predicts pixels directly. The authors observe that using pixels as targets is just as effective as using tokenized targets, but without the complexity of an extra tokenizer or additional data. Overall, the methodology is really simple: a single Transformer tries to paint in the missing pieces of an image. Despite (or rather thanks to) this simplicity, MAE scales greatly and learns rich visual representations, as we discuss next.

\begin{figure}[!b]
    \centering
    \includegraphics[width=1\linewidth]{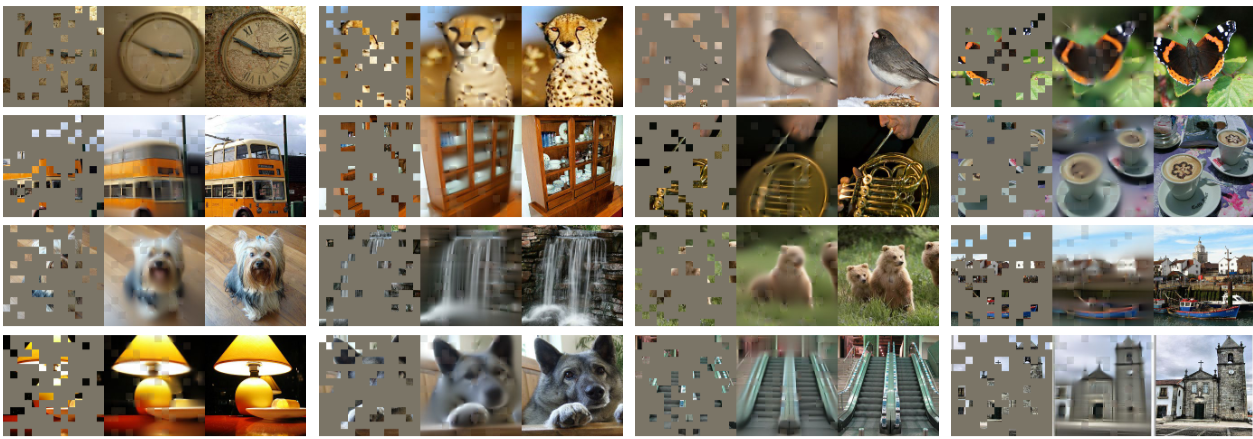}
    \caption{Example results on ImageNet validation images. For each triplet, the masked image (left), the MAE reconstruction$^\dagger$ (middle), and the ground-truth (right) are shown. The masking ratio is 80\%, leaving only 39 out of 196 patches.
    \\\textit{$^\dagger$ As no loss is computed on visible patches, the model output on visible patches is qualitatively worse. One can simply overlay the output with the visible patches to improve visual quality. Authors intentionally opt not to do this, so the method’s behavior can be understood more comprehensively.}}
    \label{fig:mae-recon}
\end{figure}

\subsubsection{Experiments and Results}

\paragraph{Image Classification (ImageNet-1K):} MAE pre-training produces competitive features for image recognition. After self-supervised pre-training on ImageNet, the encoder is fine-tuned on labeled ImageNet-1K and achieves excellent accuracy. For instance, a ViT-H model pre-trained with MAE for 1600 epochs reaches 87.8\% top-1 accuracy on ImageNet-1K. This is the highest reported accuracy for a method using only ImageNet-1K data, outperforming previous self-supervised approaches (the prior best was 87.1\%) and even matching or surpassing some methods that use larger external datasets. Importantly, even smaller models benefit: with ViT-B/16 (Base size), MAE fine-tunes to $\sim83.6\%$, slightly above contemporaneous methods like DINO~\cite{caron2021emerging} (82.8\% on ViT-B) and MoCo-v3~\cite{chen2021empirical}. The advantage grows with model size: MAE with ViT-L (85.9\%) exceeds MoCo-v3’s 84.1\% on ViT-L, indicating better scalability. Figure~\ref{fig:mae-recon} qualitatively shows MAE reconstructions: Even with 75-80\% of patches masked, the decoder can predict globally coherent images, often capturing the semantics of the scene. These reconstructions are not always pixel-perfect as they may be blurry or differ in fine details, but they are usually statistically and semantically plausible. Such results show that the MAE pre-training task indeed forces the model to learn meaningful high-level visual features, since purely low-level completion would not work for large missing regions. 

\paragraph{High Masking Ratio and Efficiency:} Experiments confirm that a high masking ratio (around 75\%) is optimal for MAE. Masking too little (e.g., 25\%) leaves an easy task that the network can solve using local correlations, yielding poorer features; masking much more than 75\% becomes extremely difficult to learn. At $\sim75\%$ masked, both fine-tune accuracy and even linear probing accuracy are maximized in the ablations. An important consequence is efficiency: with 75\% masks, the encoder processes only 25\% of the patch tokens from each image. Combined with the lightweight decoder, this leads to much faster training. MAE can be trained for 1600 epochs and still use less overall compute than contrastive methods trained for far fewer epochs. For example, using the same hardware, a ViT-L model trained with MAE for 1600 epochs took 31 hours, whereas MoCo-v3 took 36 hours for just 300 epochs. The efficiency gains come from the reduced token count per image and allow MAE to scale to longer training schedules and larger models without extreme cost. 

\paragraph{Transfer Learning and Other Tasks:} The learned representations from MAE prove effective beyond ImageNet classification. He \textit{et al.} evaluate MAE pre-trained models on downstream tasks like object detection and semantic segmentation. In semantic segmentation on ADE20K, for instance, an MAE-pretrained ViT-L yields higher mIoU than the same model with supervised ImageNet pre-training, indicating better transfer quality. This is attributed to MAE learning more generalized features that are not overfit to the ImageNet classification task. 

\paragraph{}In summary, the experiments validate MAE’s design: it yields state-of-the-art results on ImageNet and strong transfer performance, all with a simple training scheme that is efficient and scalable.

\subsubsection{Discussion and Conclusion}

\paragraph{}MAE demonstrates that masked autoencoding can be a powerful paradigm for vision self-supervised learning when designed appropriately. The approach’s simplicity is a key strength: there is no need for complex contrastive losses, negative pairs, momentum teachers, or pretext-specific heuristics. Instead, the task is conceptually straightforward: hide a large portion of the image and predict it. This simplicity leads to an extremely effective learning signal. MAE simply leverages the ViT architecture (which, unlike CNNs, naturally handles token masking and lacks strong locality bias), an asymmetric encoder-decoder that makes training on high mask ratios feasible, and a high masking ratio to induce learning of global structure efficiently. In essence, MAE is performing high-level inpainting; to fill in 75\% of missing content, the model must understand context, object shapes, and semantic cues from the visible patches, rather than just colorizing textures. This leads to representations that capture overall semantics. 

\paragraph{}Compared to other contrastive, and self distillation methods such as MoCo \cite{he2020momentum} and DINO\cite{caron2021emerging}, MAE follows a reconstructive route. Contrastive approaches learn by discriminating between different images or views, encouraging invariance to augmentations and yielding features that are immediately "clusterable" or semantically separated. Self-distillation aproaches like DINO teach a student model to replicate the outputs of a momentum‑updated teacher on several image crops, removing the need for explicit negative pairs and placing them outside the contrastive family. Generative reconstruction, on the other hand, does not explicitly enforce the separability of image instances but forces the model to model the distribution of pixel content. As a result, MAE’s latent features are slightly less separable without fine-tuning, but they contain rich information extractable with light supervision. After full fine-tuning, MAE-pretrained models outperform contrastive pre-training methods on classification tasks, and they continue to improve with larger model size, a sign of better scaling. In fact, MAE is shown to have the ability to train very large ViTs on ImageNet-1K from scratch, which previously would overfit without many extra labeled images. This scaling behavior mirrors NLP, where larger masked-language models only get better. 

\paragraph{}Another relevant point is the reconstruction target. MAE’s strong results using raw pixel targets challenge the assumption that one must predict high-level features or semantic tokens to learn semantics. The results suggest that with enough masking, predicting pixels is not a low-level task, but rather forces learning of high-level concepts (because easy low-level cues are removed). This simplifies the pipeline by removing the need for an external tokenizer or semantic labels during pre-training. It also means MAE directly optimizes in the input space, making it conceptually aligned with classic denoising autoencoders, though on a much larger scale. 

\paragraph{}In conclusion, Masked Autoencoders offer a simple, scalable framework for self-supervised learning that achieves excellent results. MAE captures meaningful visual representations without labels by cleverly masking the input and by leveraging an efficient transformer-based design. Its success highlights that generative/reconstructive pre-text tasks, long thought to be inferior to contrastive ones for vision, can perform greatly when the task difficulty and architecture are right. The MAE approach has inspired further research in masked image modeling, hybrid self-supervised techniques (combining reconstruction with contrastive objectives), and scaling up vision models. It stands as an elegant demonstration that sometimes the easiest-to-implement ideas (in this case, hide-and-predict) can yield state-of-the-art performance when executed properly.

\newpage
\section{Conclusions}
The trajectory of progress shown by these six papers reveals a clear and cohesive narrative in computer vision research. We observe a foundational shift from specialized architectures, like the convolution-centric ResNet, to the more generalized, sequence-based Vision Transformer, demonstrating that powerful visual representations can be learned with fewer inductive biases when sufficient data is available. This evolution in architectures has been coupled with advancements in their application. Generative modeling has progressed from the game-theoretic min-max principle of GANs to the more stable and computationally efficient Latent Diffusion Models, which achieve state-of-the-art synthesis by operating in a compressed latent space. Critically, the scalability and data hunger of these large models are addressed by self-supervised learning. Methods like DINO and MAE have proven essential, demonstrating that by defining difficult and creative pretext tasks such as self-distillation or reconstructing heavily masked inputs, it is possible to learn robust, semantic-rich features without reliance on human-provided labels. These paradigms are not independent but are converging: the future points toward the development of massive, self-supervised foundation models, likely Transformer-based, that serve as a unified backbone for a wide array of downstream applications, including both high-level recognition and high-fidelity, controllable generation. 
%------------------------------------------------------------------------- 
% \printbibliography[heading=subbibliography]
\bibliographystyle{plain} % or ieee, acm, etc. depending 
\bibliography{latex8.bib}

\begin{thebibliography}{10}

\bibitem{alaparthi2021bert}
Shivaji Alaparthi and Manit Mishra.
\newblock Bert: A sentiment analysis odyssey.
\newblock {\em Journal of Marketing Analytics}, 9(2):118--126, 2021.

\bibitem{bao2021beit}
Hangbo Bao, Li~Dong, Songhao Piao, and Furu Wei.
\newblock Beit: Bert pre-training of image transformers.
\newblock {\em arXiv preprint arXiv:2106.08254}, 2021.

\bibitem{brock2019largescalegantraining}
Andrew Brock, Jeff Donahue, and Karen Simonyan.
\newblock Large scale gan training for high fidelity natural image synthesis, 2019.

\bibitem{caron2021emerging}
Mathilde Caron, Hugo Touvron, Ishan Misra, Herv{\'e} J{\'e}gou, Julien Mairal, Piotr Bojanowski, and Armand Joulin.
\newblock Emerging properties in self-supervised vision transformers.
\newblock In {\em Proceedings of the IEEE/CVF international conference on computer vision}, pages 9650--9660, 2021.

\bibitem{chen2020simpleframeworkcontrastivelearning}
Ting Chen, Simon Kornblith, Mohammad Norouzi, and Geoffrey Hinton.
\newblock A simple framework for contrastive learning of visual representations, 2020.

\bibitem{chen2021empirical}
Xinlei Chen, Saining Xie, and Kaiming He.
\newblock An empirical study of training self-supervised vision transformers.
\newblock In {\em Proceedings of the IEEE/CVF international conference on computer vision}, pages 9640--9649, 2021.

\bibitem{deininger2022comparative}
Luca Deininger, Bernhard Stimpel, Anil Yuce, Samaneh Abbasi-Sureshjani, Simon Sch{\"o}nenberger, Paolo Ocampo, Konstanty Korski, and Fabien Gaire.
\newblock A comparative study between vision transformers and cnns in digital pathology.
\newblock {\em arXiv preprint arXiv:2206.00389}, 2022.

\bibitem{devlin-etal-2019-bert}
Jacob Devlin, Ming-Wei Chang, Kenton Lee, and Kristina Toutanova.
\newblock {BERT}: Pre-training of deep bidirectional transformers for language understanding.
\newblock In Jill Burstein, Christy Doran, and Thamar Solorio, editors, {\em Proceedings of the 2019 Conference of the North {A}merican Chapter of the Association for Computational Linguistics: Human Language Technologies, Volume 1 (Long and Short Papers)}, pages 4171--4186, Minneapolis, Minnesota, June 2019. Association for Computational Linguistics.

\bibitem{dhiwise2025gantransformer}
DhiWise.
\newblock Gan vs transformer: A generative ai comparison, June 2025.
\newblock Accessed: 2025-07-05.

\bibitem{dosovitskiy2020image}
Alexey Dosovitskiy, Lucas Beyer, Alexander Kolesnikov, Dirk Weissenborn, Xiaohua Zhai, Thomas Unterthiner, Mostafa Dehghani, Matthias Minderer, Georg Heigold, Sylvain Gelly, et~al.
\newblock An image is worth 16x16 words: Transformers for image recognition at scale.
\newblock {\em arXiv preprint arXiv:2010.11929}, 2020.

\bibitem{goodfellow2014generative}
Ian~J Goodfellow, Jean Pouget-Abadie, Mehdi Mirza, Bing Xu, David Warde-Farley, Sherjil Ozair, Aaron Courville, and Yoshua Bengio.
\newblock Generative adversarial nets.
\newblock {\em Advances in neural information processing systems}, 27, 2014.

\bibitem{he2022masked}
Kaiming He, Xinlei Chen, Saining Xie, Yanghao Li, Piotr Doll{\'a}r, and Ross Girshick.
\newblock Masked autoencoders are scalable vision learners.
\newblock In {\em Proceedings of the IEEE/CVF conference on computer vision and pattern recognition}, pages 16000--16009, 2022.

\bibitem{he2020momentum}
Kaiming He, Haoqi Fan, Yuxin Wu, Saining Xie, and Ross Girshick.
\newblock Momentum contrast for unsupervised visual representation learning.
\newblock In {\em Proceedings of the IEEE/CVF conference on computer vision and pattern recognition}, pages 9729--9738, 2020.

\bibitem{he2016deep}
Kaiming He, Xiangyu Zhang, Shaoqing Ren, and Jian Sun.
\newblock Deep residual learning for image recognition.
\newblock In {\em Proceedings of the IEEE conference on computer vision and pattern recognition}, pages 770--778, 2016.

\bibitem{heusel2018ganstrainedtimescaleupdate}
Martin Heusel, Hubert Ramsauer, Thomas Unterthiner, Bernhard Nessler, and Sepp Hochreiter.
\newblock Gans trained by a two time-scale update rule converge to a local nash equilibrium, 2018.

\bibitem{ho2020denoising}
Jonathan Ho, Ajay Jain, and Pieter Abbeel.
\newblock Denoising diffusion probabilistic models, 2020.

\bibitem{lstm}
Sepp Hochreiter and J\"{u}rgen Schmidhuber.
\newblock Long short-term memory.
\newblock {\em Neural Comput.}, 9(8):1735–1780, November 1997.

\bibitem{karras2018progressivegrowinggansimproved}
Tero Karras, Timo Aila, Samuli Laine, and Jaakko Lehtinen.
\newblock Progressive growing of gans for improved quality, stability, and variation, 2018.

\bibitem{kolesnikov2020big}
Alexander Kolesnikov, Lucas Beyer, Xiaohua Zhai, Joan Puigcerver, Jessica Yung, Sylvain Gelly, and Neil Houlsby.
\newblock Big transfer (bit): General visual representation learning.
\newblock In {\em European conference on computer vision}, pages 491--507. Springer, 2020.

\bibitem{krizhevsky2012imagenet}
Alex Krizhevsky, Ilya Sutskever, and Geoffrey~E Hinton.
\newblock Imagenet classification with deep convolutional neural networks.
\newblock {\em Advances in neural information processing systems}, 25, 2012.

\bibitem{li2022efficientselfsupervisedvisiontransformers}
Chunyuan Li, Jianwei Yang, Pengchuan Zhang, Mei Gao, Bin Xiao, Xiyang Dai, Lu~Yuan, and Jianfeng Gao.
\newblock Efficient self-supervised vision transformers for representation learning, 2022.

\bibitem{aiblog}
Allison Linn.
\newblock {Microsoft researchers win ImageNet computer vision challenge}.
\newblock \url{https://blogs.microsoft.com/ai/microsoft-researchers-win-imagenet-computer-vision-challenge/}.
\newblock Accessed: 2025-07-16.

\bibitem{liu2021swin}
Ze~Liu, Yutong Lin, Yue Cao, Han Hu, Yixuan Wei, Zheng Zhang, Stephen Lin, and Baining Guo.
\newblock Swin transformer: Hierarchical vision transformer using shifted windows.
\newblock In {\em Proceedings of the IEEE/CVF international conference on computer vision}, pages 10012--10022, 2021.

\bibitem{openai_chatgpt}
{OpenAI}.
\newblock {ChatGPT}, 2025.
\newblock Large language model, accessed 2025-07-20.

\bibitem{oquab2024dinov2learningrobustvisual}
Maxime Oquab, Timothée Darcet, Théo Moutakanni, Huy Vo, Marc Szafraniec, Vasil Khalidov, Pierre Fernandez, Daniel Haziza, Francisco Massa, Alaaeldin El-Nouby, Mahmoud Assran, Nicolas Ballas, Wojciech Galuba, Russell Howes, Po-Yao Huang, Shang-Wen Li, Ishan Misra, Michael Rabbat, Vasu Sharma, Gabriel Synnaeve, Hu~Xu, Hervé Jegou, Julien Mairal, Patrick Labatut, Armand Joulin, and Piotr Bojanowski.
\newblock Dinov2: Learning robust visual features without supervision, 2024.

\bibitem{rombach2022high}
Robin Rombach, Andreas Blattmann, Dominik Lorenz, Patrick Esser, and Björn Ommer.
\newblock High-resolution image synthesis with latent diffusion models, 2022.

\bibitem{shamsolmoali2020imagesynthesisadversarialnetworks}
Pourya Shamsolmoali, Masoumeh Zareapoor, Eric Granger, Huiyu Zhou, Ruili Wang, M.~Emre Celebi, and Jie Yang.
\newblock Image synthesis with adversarial networks: a comprehensive survey and case studies, 2020.

\bibitem{Simonyan2014}
Karen Simonyan and Andrew Zisserman.
\newblock Very deep convolutional networks for large-scale image recognition.
\newblock September 2014.

\bibitem{highway_networks}
Rupesh~Kumar Srivastava, Klaus Greff, and J{\"{u}}rgen Schmidhuber.
\newblock Highway networks.
\newblock {\em CoRR}, abs/1505.00387, 2015.

\bibitem{szegedy2014goingdeeperconvolutions}
Christian Szegedy, Wei Liu, Yangqing Jia, Pierre Sermanet, Scott Reed, Dragomir Anguelov, Dumitru Erhan, Vincent Vanhoucke, and Andrew Rabinovich.
\newblock Going deeper with convolutions, 2014.

\bibitem{tuli2021convolutional}
Shikhar Tuli, Ishita Dasgupta, Erin Grant, and Thomas~L Griffiths.
\newblock Are convolutional neural networks or transformers more like human vision?
\newblock {\em arXiv preprint arXiv:2105.07197}, 2021.

\bibitem{vaswani_2017_attention}
Ashish Vaswani, Google Brain, Noam Shazeer, Niki Parmar, Jakob Uszkoreit, Llion Jones, Aidan Gomez, Łukasz Kaiser, and Illia Polosukhin.
\newblock Attention is all you need, 06 2017.

\bibitem{wang2018non}
Xiaolong Wang, Ross Girshick, Abhinav Gupta, and Kaiming He.
\newblock Non-local neural networks.
\newblock In {\em Proceedings of the IEEE conference on computer vision and pattern recognition}, pages 7794--7803, 2018.

\bibitem{zhai2022scaling}
Xiaohua Zhai, Alexander Kolesnikov, Neil Houlsby, and Lucas Beyer.
\newblock Scaling vision transformers.
\newblock In {\em Proceedings of the IEEE/CVF conference on computer vision and pattern recognition}, pages 12104--12113, 2022.

\bibitem{zhang2019progressiveaugmentationgans}
Dan Zhang and Anna Khoreva.
\newblock Progressive augmentation of gans, 2019.

\bibitem{zhao2017energybasedgenerativeadversarialnetwork}
Junbo Zhao, Michael Mathieu, and Yann LeCun.
\newblock Energy-based generative adversarial network, 2017.

\bibitem{zhou2022ibotimagebertpretraining}
Jinghao Zhou, Chen Wei, Huiyu Wang, Wei Shen, Cihang Xie, Alan Yuille, and Tao Kong.
\newblock ibot: Image bert pre-training with online tokenizer, 2022.

\end{thebibliography}

\end{document}